\documentclass[sn-standardnature]{sn-jnl}

\jyear{2021}%

\theoremstyle{thmstyleone}%
\theoremstyle{thmstyletwo}%

\theoremstyle{thmstylethree}%

\usepackage{color,soul}
\newenvironment{grey}{\color{lightgray}}{\ignorespacesafterend}

\usepackage{bbm}
\usepackage{nicefrac}

\usepackage{graphicx}
\usepackage{rotating}
\usepackage{subcaption}
\usepackage{placeins}

\usepackage{booktabs}
\usepackage{multirow}
\usepackage{multicol}
\usepackage{makecell}
\usepackage{threeparttable}
\usepackage{verbatimbox}
\usepackage{tabularx}
\usepackage{longtable}

\begin{document}

\title[Generalisability of DL for prediction in the ICU]{From Single-Hospital to Multi-Centre Applications: Enhancing the Generalisability of Deep Learning Models for Adverse Event Prediction in the ICU}


\author*[1,2]{\fnm{Patrick} \sur{Rockenschaub}}\email{rockenschaub.patrick@gmail.com}

\author[1]{\fnm{Adam} \sur{Hilbert}}\email{adam.hilbert@charite.de}

\author[1]{\fnm{Tabea} \sur{Kossen}}\email{tabea.kossen@charite.de}

\author[3]{\fnm{Falk} \sur{von Dincklage}}\email{falk.vondincklage@med.uni-greifswald.de}

\author[2,4]{\fnm{Vince} \fnm{Istvan} \sur{Madai}}\email{vince\_istvan.madai@bih-charite.de}

\author[1]{\fnm{Dietmar} \sur{Frey}}\email{dietmar.frey@charite.de}

\affil*[1]{\orgdiv{Charité Lab for Artificial Intelligence in Medicine — CLAIM}, \orgname{Charité - Universitätsmedizin Berlin}, \orgaddress{\street{Virchowweg 10}, \city{Berlin}, \postcode{10117}, \country{Germany}}}

\affil[2]{\orgdiv{QUEST Center for Responsible Research}, \orgname{Berlin Institute of Health at Charité – Universitätsmedizin Berlin}, \orgaddress{\street{Anna-Louisa-Karsch-Straße 2}, \city{Berlin}, \postcode{10178}, \country{Germany}}}

\affil[3]{\orgdiv{Department of Anesthesia, Intensive Care, Emergency and Pain Medicine}, \orgname{Universitätsmedizin Greifswald}, \orgaddress{\street{Ferdinand-Sauerbruch-Straße}, \city{Greifswald}, \postcode{17475}, \country{Germany}}}

\affil[4]{\orgdiv{Faculty of Computing, Engineering and the Built Environment, School of Computing and Digital Technology}, \orgname{Birmingham City University}, \orgaddress{\street{Curzon Street}, \city{Birmingham}, \postcode{B4 7XG}, \country{United Kingdom}}}


\abstract{Deep learning (DL) can aid doctors in detecting worsening patient states early, affording them time to react and prevent bad outcomes. While DL-based early warning models usually work well in the hospitals they were trained for, they tend to be less reliable when applied at new hospitals. This makes it difficult to deploy them at scale. Using carefully harmonised intensive care data from four data sources across Europe and the US (totalling 334,812 stays), we systematically assessed the reliability of DL models for three common adverse events: death, acute kidney injury (AKI), and sepsis. We tested whether using more than one data source and/or explicitly optimising for generalisability during training improves model performance at new hospitals. We found that models achieved high AUROC for mortality (0.838-0.869), AKI (0.823-0.866), and sepsis (0.749-0.824) at the training hospital. As expected, performance dropped at new hospitals, sometimes by as much as -0.200. Using more than one data source for training mitigated the performance drop, with multi-source models performing roughly on par with the best single-source model. This suggests that as data from more hospitals become available for training, model robustness is likely to increase, lower-bounding robustness with the performance of the most applicable data source in the training data. Dedicated methods promoting generalisability did not noticeably improve performance in our experiments. }

\keywords{machine learning, intensive care, electronic health records, prediction, generalisability}



\maketitle

\section{Introduction}\label{sec1}

The increasing availability of routine electronic health records has led to a burgeoning of deep learning (DL)-based clinical decision support systems \cite{kellyKeyChallengesDelivering2019}. Both academic groups and start-ups promise to harness the wealth of medical data to create tools that aid doctors and improve patient outcomes. Applications in intensive care units (ICUs) are especially prevalent, owing to the need for close monitoring in this setting and a consequently high density of (digitised) measurements \cite{shillanUseMachineLearning2019a}. To date, prediction  models have been proposed for a range of ICU events including ICU mortality \cite{silvaPredictingInHospitalMortality2012, pirracchioMortalityPredictionIntensive2015a}, acute kidney injury (AKI) \cite{meyerMachineLearningRealtime2018, koynerDevelopmentMachineLearning2018a}, and sepsis \cite{reynaEarlyPredictionSepsis2020, moorPredictingSepsisMultisite2021}. 

A major challenge for the translation of these models into clinical practice is that most were trained on data from a single hospital. Multi-centre studies remain scarce due to continuing difficulties of accessing and harmonising data \cite{kellyKeyChallengesDelivering2019}. Existing models thus implicitly assume that future patients come from a hospital \textit{just like the one} seen during training \cite{gulrajaniSearchLostDomain2020}. Yet, hospitals differ in terms of the patients that they see or the clinical pathways in diagnostics and treatment. \cite{sauerSystematicReviewComparison2022a}. A DL model may exploit such local patterns of care \cite{futomaGeneralizationClinicalPrediction2021}. For example, a local sepsis treatment pathway may prescribe a specific set of investigations when there is suspicion of infection. Models that merely learn to look for these investigations --- rather than the underlying biology of sepsis --- may perform well at the training hospital but will do worse at new hospitals with different pathways \cite{reynaEarlyPredictionSepsis2020}. Deploying such models at scale will require extensive re-training and re-testing at every new site. 

Intuitively, training on data from more than one hospital may alleviate this \cite{kellyKeyChallengesDelivering2019}. Associations that are present at multiple hospitals will be more likely to apply to new hospitals too \cite{gulrajaniSearchLostDomain2020}. However, little is known on how effective this approach is in practice \cite{wynantsUntappedPotentialMulticenter2019}, especially for the often messy, high-dimensional data collected in ICUs. Flexible DL models may additionally require training algorithms that explicitly optimise for generalisability \cite{sunDeepCORALCorrelation2016a, kruegerOutofDistributionGeneralizationRisk2020, rameFishrInvariantGradient2021, liLearningGeneralizeMetaLearning2017, sagawaDistributionallyRobustNeural2019}. Better understanding how we can improve the generalisability of DL models for the ICU is crucial to developing trustworthy applications that have a real chance of being translated into clinical practice. 

In this work, we used carefully harmonised data from four ICU data sources across the US and Europe to investigate how reliably DL models can be transferred from one hospital to another. We provide results for three common ICU prediction tasks: ICU mortality, AKI, and sepsis. Using these tasks, we study if training on multiple data sources improves model performance at new hospitals and quantify any benefits derived from explicitly optimising for generalisability. 

\section{Results}\label{sec2}

\begin{table}
    \footnotesize
    \caption{Patient characteristics.}
    \label{tab:tableone}
    \centering
    \begin{threeparttable}
        \begin{tabularx}{\textwidth}{Xcccc}
            \toprule
             & \textbf{AUMCdb} & \textbf{HiRID} & \textbf{eICU} & \textbf{MIMIC IV} \\
            \midrule
            \addlinespace[0.3em]
            \textbf{Number of patients} & 19,790 & -\footnotemark[1] & 160,816 & 53,090 \\
            \textbf{Number of ICU stays} & 20,636 & 32,338 & 182,774 & 75,652  \\
            \addlinespace[0.6em]
            \textbf{Age at admission} (years) & 65 [55, 75]\footnotemark[2] & 65 [55, 75] & 65 [53, 76] & 65 [53, 76] \\
            \textbf{Female} & 7,699 (35) & 11,542 (36) & 83,940 (46) & 33,499 (44) \\
            \multicolumn{5}{l}{\textbf{Race}}\\
            \hspace{1em}Asian & - & - & 3,008 (3) & 2,225 (3) 	\\
            \hspace{1em}Black & - & - & 19,867 (11) & 8,223 (12) \\
            \hspace{1em}White & - & - & 140,938 (78) & 51,575 (76) \\
            \hspace{1em}Other & - & - & 16,978 (9) & 5,514 (8)	\\
            \hspace{1em}Unknown & & & 1,983 & 8,115 \\
            \addlinespace[0.5em]
            \multicolumn{5}{l}{\textbf{Admission type}}\\
            \hspace{1em}Medical & 4,131 (21) & - & 134,532 (79) & 49,217 (65) \\
            \hspace{1em}Surgical & 14,007 (72) & - & 31,909 (19) & 25,674 (34)\\
            \hspace{1em}Other & 1,225 (6) & - & 4,702 (3) & 761 (1)	\\
            \hspace{1em}Unknown & 1,069 & - & 11,631 & 0 \\
            \addlinespace[0.5em]
            \textbf{ICU length of stay} (hours) & 24 [19, 77] & 24 [19, 50] & 42 [23, 76] & 48 [26, 89] \\
            \textbf{Hospital length of stay} (days) & - & - & 6 [3, 10] & 7 [4, 13] \\
            \midrule
            \multicolumn{5}{l}{\textbf{ICU mortality}} \\
            \hspace{1em}Number of included stays & 10,535 & 12,859 & 113,382 & 52,045 \\
            \hspace{1em}Died & 1,660 (15.8) & 1,097 (8.2) & 6,253 (5.5) & 3,779 (7.3) \\
            \multicolumn{5}{l}{\textbf{Onset of AKI}}\\
            \hspace{1em}Number of included stays & 20,290 & 31,772 & 164,791 & 66,032\\
            \hspace{1em}KDIGO $\geq$ 1 & 3,776 (18.6) & 7,383 (23.2) & 62,535 (37.9) & 27,509 (41.7) \\
            \multicolumn{5}{l}{\textbf{Onset of Sepsis}}\\
            \hspace{1em}Number of included stays & 18,184 & 29,894 & 123,864 & 67,056 \\
            \hspace{1em}Sepsis-3 criteria & 764 (4.2) & 1,986 (6.6) & 5,835 (4.7) & 3,730 (5.6) \\
            \bottomrule
        \end{tabularx}
        \begin{tablenotes}[flushleft, para]
            \footnotesize
            \item \leavevmode\kern-\scriptspace\kern-\labelsep
            \footnotemark[1] HiRID only provides stay-level identifiers.
            \footnotemark[2] Since AUMCdb only includes age groups, we calculated the median of the group midpoints.
            \item \leavevmode\kern-\scriptspace\kern-\labelsep Numeric variables are summarised by median [IQR] and categorical variables are summarised by N (\%). AKI, acute kidney injury; ICU, intensive care unit; KDIGO, Kidney Disease Improving Global Outcomes.
        \end{tablenotes}
    \end{threeparttable}
\end{table}

\begin{figure}
\centering
\begin{subfigure}{.45\textwidth}
    \caption{\textbf{ICU mortality}}
    \includegraphics[scale=0.5]{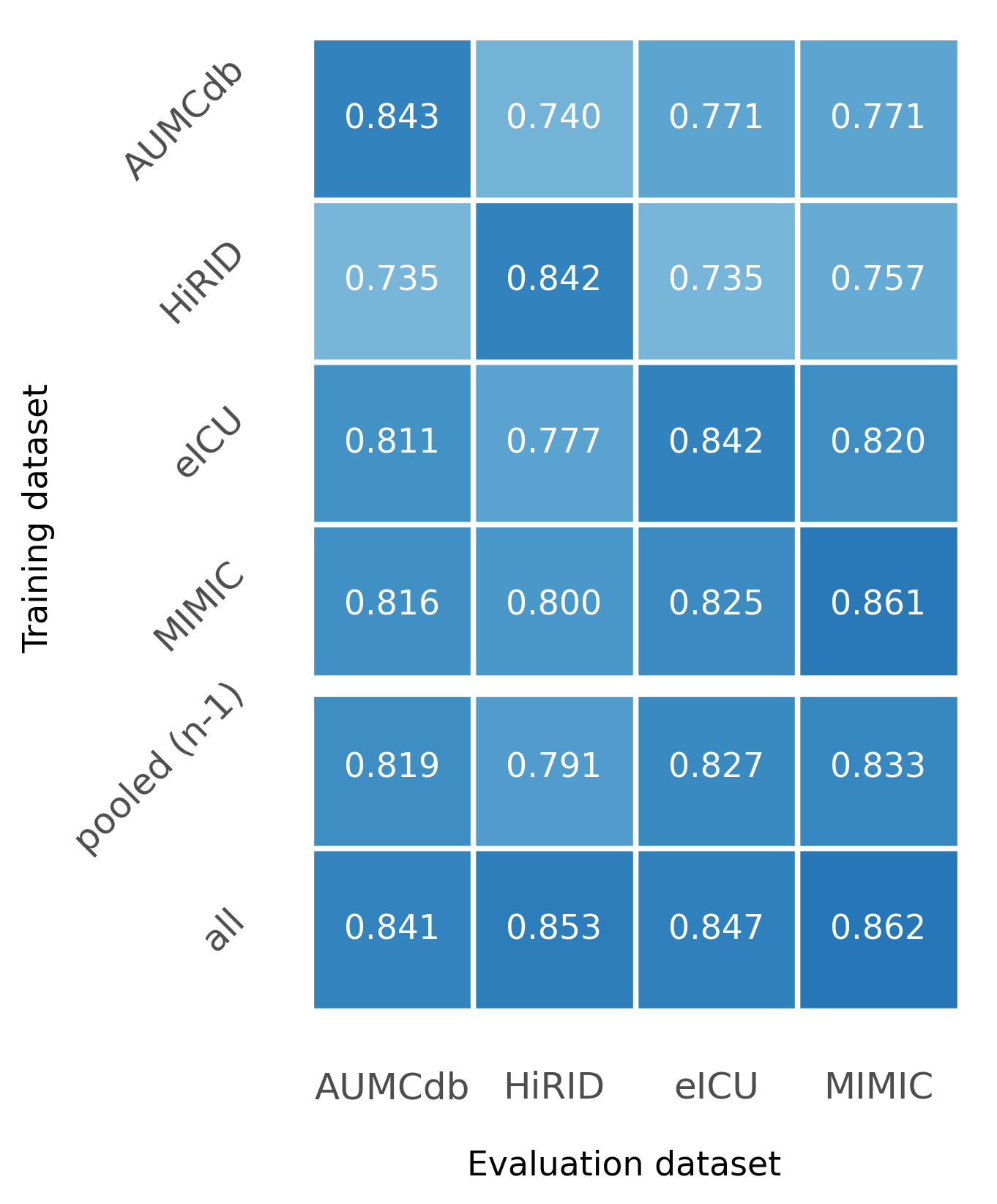}
\end{subfigure}\hfill%
\begin{subfigure}{.45\textwidth}
    \centering
    \caption{\textbf{AKI}}
    \includegraphics[scale=0.5]{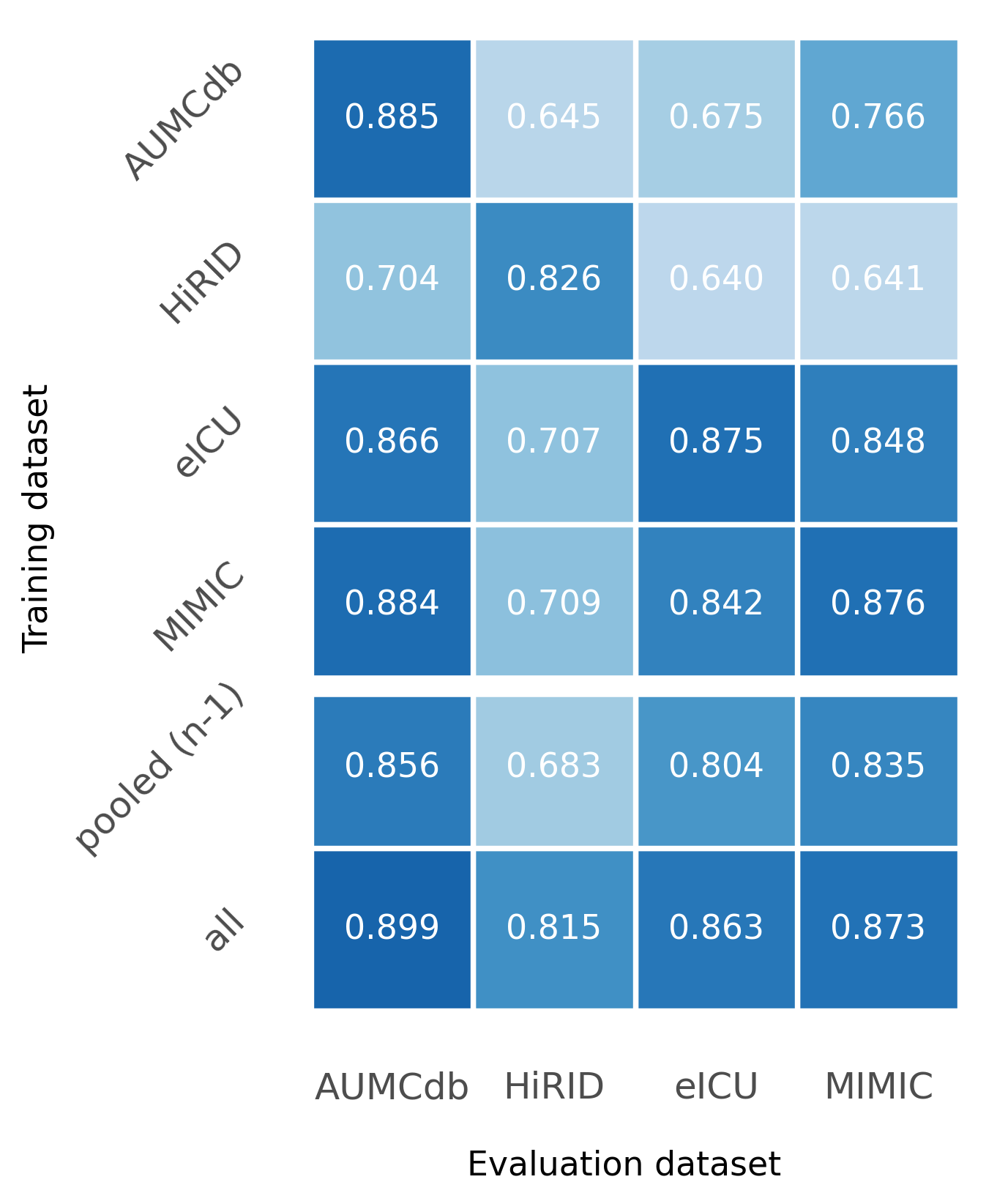}
\end{subfigure}
\begin{subfigure}{.45\textwidth}
    \centering
    \caption{\textbf{Sepsis}}
    \includegraphics[scale=0.5]{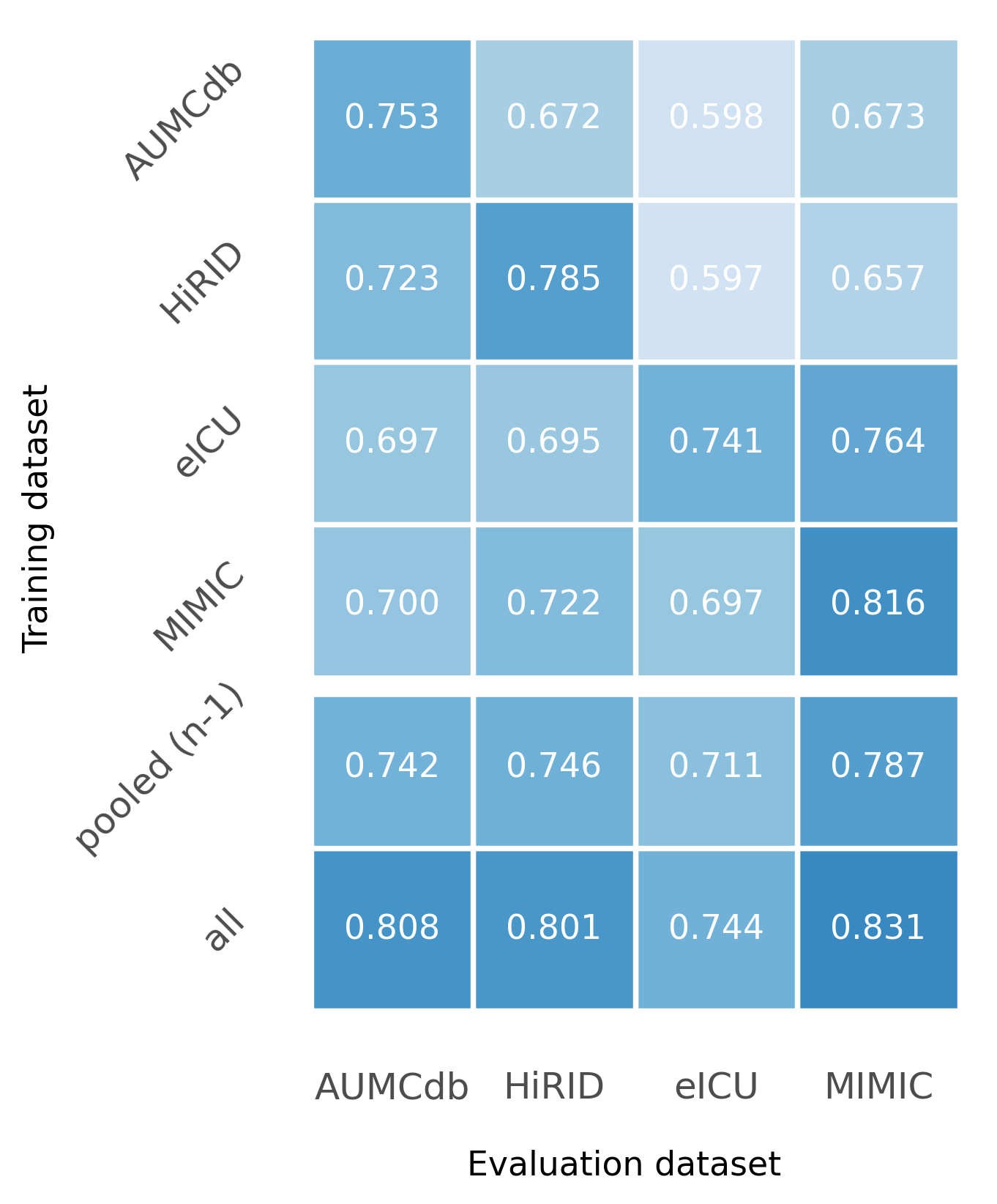}
\end{subfigure}\hfill%
\begin{subfigure}{.45\textwidth}
    \includegraphics[scale=0.5]{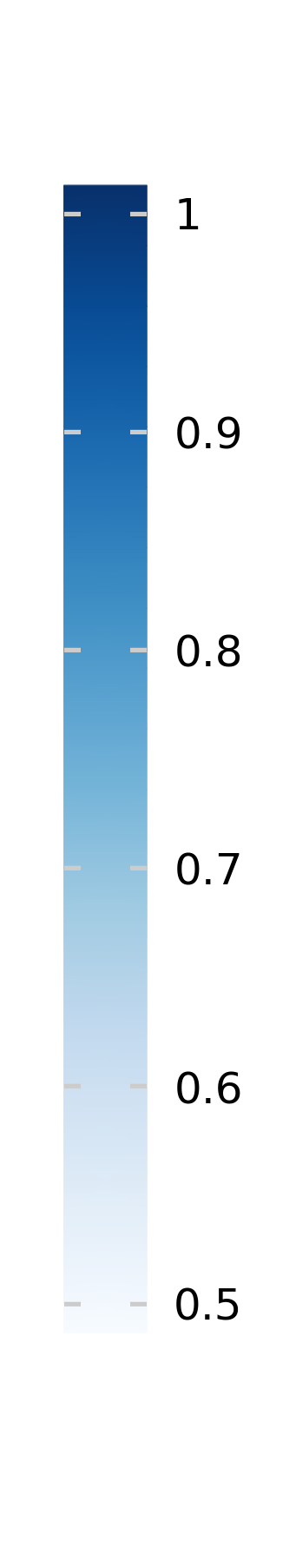}
\end{subfigure}
\caption[short]{AUROC achieved by a) ICU mortality, b) AKI, and c) sepsis prediction models when trained on one dataset (rows) and evaluated on others (columns). The diagonal represents in-dataset validation, i.e., training and test samples were taken from the same dataset. In \textit{pooled (n-1)}, the model was trained on combined data from all except the test dataset. In \textit{all}, training data from all datasets (including the test dataset) was used during model development. Models used a GRU featuriser. \footnotesize AKI, acute kidney injury; AUROC, area under the receiver operating characteristic; ICU, intensive care unit.}
\label{fig:erm}
\end{figure}

We used data from four ICU datasets across three countries: AUMCdb (Netherlands) \cite{thoralSharingICUPatient2021}, HiRID (Switzerland) \cite{hylandEarlyPredictionCirculatory2020}, eICU (US) \cite{pollardEICUCollaborativeResearch2018, pollardEICUCollaborativeResearch2019}, and MIMIC IV (US) \cite{johnsonMIMICIV2022}. From these datasets, we combined and harmonised data for a total of 334,812 ICU stays that occurred between 2003 and 2019. After removing stays from children or from patients with insufficient data (Supplementary Figure \ref{fig:attrition-base}), we included 311,400 stays in our analysis (AUMCdb: 20,636; HiRID: 32,338; eICU: 182,774; MIMIC: 75,652). 

The median age at admission was 65 years and was the same across datasets (Table \ref{tab:tableone}). There were fewer women than men, with a more pronounced imbalance at European hospitals ($\sim$35\% women) compared to the US ($\sim$45\% women). Not every dataset provided the same information. Patients' race was only available in US datasets, where three out of four stays were from White patients. Admissions were predominantly to medical ICUs within MIMIC (65\%) and eICU (79\%). AUMCdb primarily captured admissions to surgical ICUs (72\%). No information on ICU type was available in HiRID. ICU stays in European datasets lasted for a median of 24 hours, while US stays were roughly twice as long at 48 and 42 hours (MIMIC and eICU respectively).

\subsection{Prediction tasks}

We built models for three prediction tasks: ICU mortality after 24 hours in the ICU (one prediction per stay), and AKI and sepsis onset within the next 6 hours (dynamic prediction at every hour of a stay). After additional task-specific exclusion criteria were applied (see Methods and Figure \ref{fig:attrition-tasks}), we observed 12,789 (6.8\%) deaths among 188,821 eligible stays, 101,203 (35.8\%) cases of AKI among 282,885 eligible stays, and 12,318 (5.2\%) cases of sepsis among 238,998 eligible stays. ICU mortality was lowest in eICU (5.5\%) and highest in AUMCdb (15.8\%). AKI rates ranged from 18.6\% in AUMCdb to 41.7\% in MIMIC, and were generally lower in European datasets. Sepsis rates ranged from 4.2\% in AUMCdb to 6.6\% in HiRID.

\subsection{Model performance within a single dataset}

We used deep neural networks to predict each outcome separately per data source. The models were able to predict ICU mortality well, with areas under the the receiver operating characteristic (AUROC) ranging from 0.842±0.009 in HiRID to 0.861±0.002 in MIMIC (mean ± standard error across cross-validation runs; Figure \ref{fig:erm} and Tables \ref{tab:erm-mort-detail}-\ref{tab:erm-sepsis-detail}). Discriminative performance for AKI onset was comparable, ranging from 0.826±0.003 in HiRID to 0.885±0.004 in AUMCdb. Sepsis was the most difficult task, with lower performance ranging from 0.741±0.002 in eICU to 0.816±0.004 in MIMIC. There was little difference between neural network types (Tables \ref{tab:erm-mort-detail}-\ref{tab:erm-sepsis-detail}). After re-calibrating models on the validation set via isotonic regression, calibration in the training dataset was reasonable (Figure \ref{fig:cal}), although calibrated probabilities were almost exclusively $\lt$25\% for sepsis. 

\subsection{Generalisability of prediction models to new datasets}

\begin{figure}
\includegraphics[width=\textwidth]{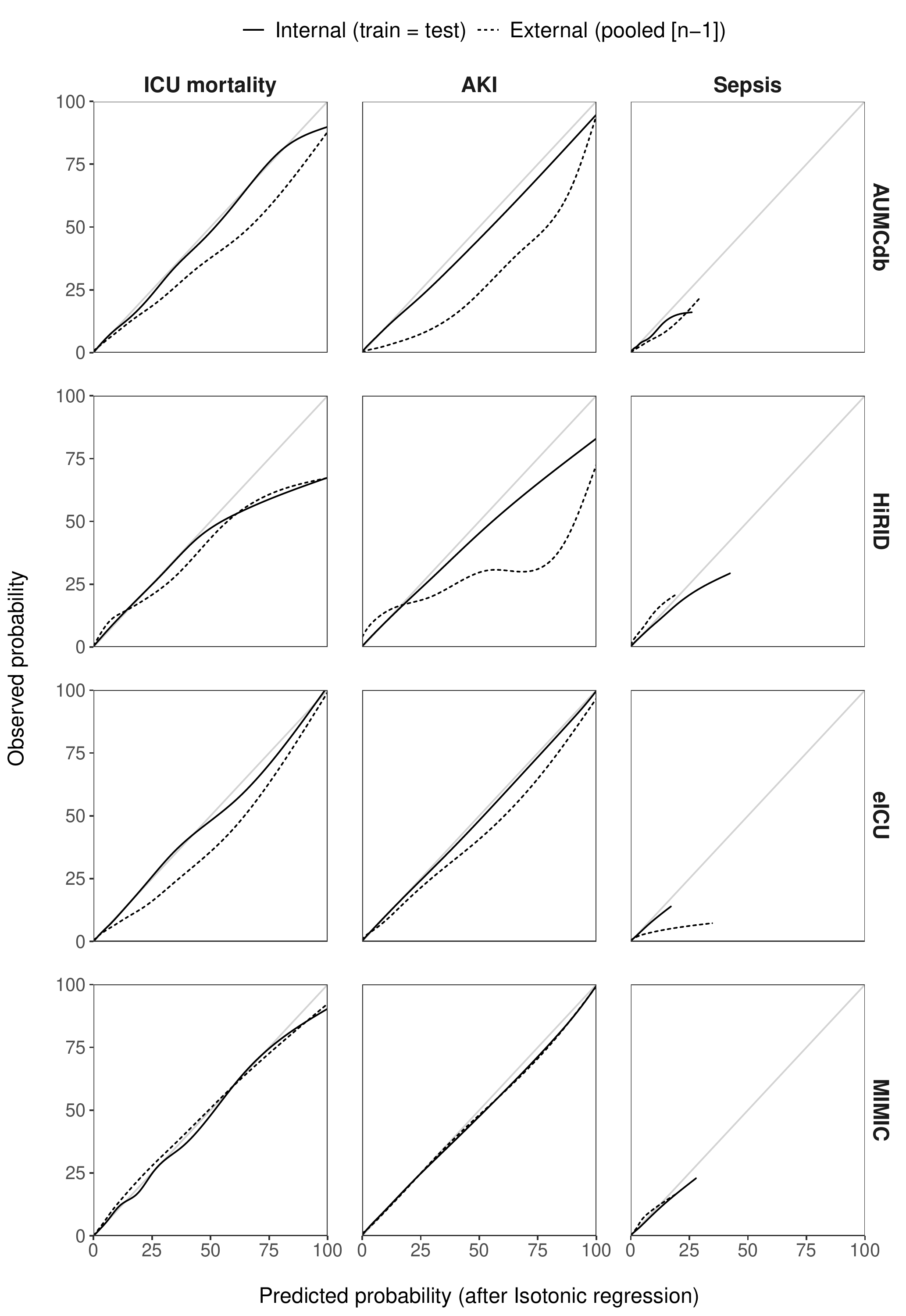}
\caption{Calibration curves by task (columns) and data source (rows). Solid lines correspond to models trained and tested on data from the same source. Dashed lines represent a model trained on data from all other data sources. For all models, the raw neural network output was calibrated on the validation split using isotonic regression. As sepsis predictions were heavily skewed with very few predicted risks $\gt 0.25$, predictions were winsorized at the 99.9th percentile to improve readability.}
\label{fig:cal}
\end{figure}

As expected, discriminative performance decreased when trained models were applied to other data sources which they hadn't seen before (Figure \ref{fig:erm} and Tables \ref{tab:erm-mort-detail}-\ref{tab:erm-sepsis-detail}) --- sometimes by as much as $\Delta_{AUROC} = 0.210$ (AKI: AUMCdb to eICU). Models trained using MIMIC or eICU tended to be more generalisable, particularly among each other. Using more than one data source for training mitigated the performance drop, with pooled models often performing on par with the best single-source model, although there were exceptions in the case of AKI. Pooled models frequently overestimated risks, though, leading to worse calibration (Figure \ref{fig:cal}). Training on all data sources simultaneously and including the test hospital(s) during training generally led to models that performed as good or better than those trained on a single source (Figure \ref{fig:erm}).

\subsection{Explicitly optimising for generalisation}

In addition to regular model training --- often called Empirical Risk Minimisation (ERM) \cite{vapnikOverviewStatisticalLearning1999} --- we evaluated five training algorithms that explicitly optimised for generalisability to new hospitals: Correlation Alignment (CORAL) \cite{sunDeepCORALCorrelation2016a}, Variance Risk Extrapolation (VREx) \cite{kruegerOutofDistributionGeneralizationRisk2020}, Fishr \cite{rameFishrInvariantGradient2021}, Meta Learning Domain Generalisation (MLDG) \cite{liLearningGeneralizeMetaLearning2017}, and Group Distributionally Robust Optimisation (GroupDRO) \cite{sagawaDistributionallyRobustNeural2019}. While the exact mechanisms by which these methods achieve their goal differ, they all promise to improve a model's generalisability to new hospitals by incentivising the model to learn relationships that can be found in more than one hospital (see Methods section and supplementary material for details). 

In our analysis, these algorithms occasionally performed better than ERM on the task of ICU mortality and AKI prediction (p-value $\lt$ 0.1) but no algorithm consistently outperformed ERM across tasks and datasets (Table \ref{tab:dg}). GroupDRO was the only algorithm that never performed worse than ERM, achieving statistically significantly higher performance in 2/12 experiments. The largest improvement from AUC 0.683±0.008 to AUC 0.754±0.002 was seen for VREx when predicting AKI on HiRID data. Notably, the same algorithm led to considerable drop in performance when similarly applied to all other data sources. 

\begin{table}[ht]
    \scriptsize
    \caption{Performance of domain generalisation algorithms compared to regular training (ERM). We format \underline{better}, same, and \begin{grey}worse\end{grey}~than ERM.}
    \label{tab:dg}
    \centering
    \begin{threeparttable}
        \begin{tabularx}{\textwidth}{Xcccc}
            \toprule
            \textbf{Algorithm} & \multicolumn{4}{c}{\textbf{Test domain}} \\\cline{2-5}
             & AUMCdb & HiRID & eICU & MIMIC \\
            \midrule
            \addlinespace[0.3em]
            \multicolumn{5}{l}{\textbf{ICU mortality}}\\
            ERM & 0.819±0.005 & 0.791±0.006 & 0.827±0.004 & 0.833±0.002 \\ 
            \addlinespace[0.5em]
            CORAL & 0.816±0.006 & 0.785±0.011 & 0.828±0.002 & 0.835±0.004 \\ 
            VREx & 0.822±0.005 & 0.791±0.007 & 0.818±0.002 & \grey{0.824±0.003} \\ 
            Fishr & 0.803±0.009 & \grey{0.679±0.006} & \grey{0.803±0.003} & \grey{0.690±0.029} \\ 
            MLDG & 0.825±0.003 & 0.801±0.003 & 0.820±0.003 & 0.837±0.001 \\ 
            GroupDRO & 0.823±0.007 & 0.801±0.008 & 0.826±0.002 & \underline{0.848±0.002} \\ 
            \midrule
            \multicolumn{5}{l}{\textbf{AKI}}\\
            ERM & 0.856±0.004 & 0.683±0.008 & 0.804±0.002 & 0.835±0.001 \\ 
            \addlinespace[0.5em]
            CORAL & 0.851±0.003 & \underline{0.713±0.003} & \grey{0.723±0.018} & \grey{0.822±0.003} \\ 
            VREx & \grey{0.802±0.003} & \underline{0.754±0.002} & \grey{0.656±0.001} & \grey{0.746±0.003} \\ 
            Fishr & \grey{0.699±0.016} & \underline{0.730±0.003} & \grey{0.636±0.011} & \grey{0.696±0.016} \\ 
            MLDG & \grey{0.840±0.006} & 0.682±0.011 & \grey{0.743±0.017} & 0.825±0.003 \\ 
            GroupDRO & 0.856±0.004 & 0.686±0.002 & \underline{0.815±0.001} & 0.833±0.001 \\ 
            \midrule
            \multicolumn{5}{l}{\textbf{Sepsis}}\\
            ERM & 0.742±0.009 & 0.746±0.003 & 0.711±0.003 & 0.787±0.004 \\ 
            \addlinespace[0.5em]
            CORAL & 0.738±0.004 & 0.744±0.003 & 0.712±0.004 & 0.783±0.003 \\ 
            VREx & \grey{0.674±0.008} & \grey{0.688±0.006} & \grey{0.672±0.004} & \grey{0.728±0.004} \\ 
            Fishr & \grey{0.562±0.026} & \grey{0.616±0.018} & \grey{0.596±0.013} & \grey{0.584±0.033} \\ 
            MLDG & 0.731±0.008 & \grey{0.726±0.005} & 0.711±0.004 & 0.778±0.004 \\ 
            GroupDRO & 0.728±0.009 & 0.734±0.003 & 0.714±0.002 & 0.785±0.004 \\ 
            \bottomrule
        \end{tabularx}
        \begin{tablenotes}[flushleft, para]
            \footnotesize
            \item \leavevmode\kern-\scriptspace\kern-\labelsep Algorithms are trained using data from all datasets except the test dataset. All models shown in this table used a GRU featuriser. Performances were compared to that of ERM using a paired Wilcoxon rank-sum test across folds. Underscored (greyed-out) values indicate statistically significantly better (worse) performance at p$<$0.1. AKI, acute kidney injury; ERM, empirical risk minimisation; CORAL; correlation alignment; GroupDRO; group distributionally robust optimisation; MLDG; meta learning domain generalisation; VREx; variance risk extrapolation.
        \end{tablenotes}
    \end{threeparttable}
\end{table}

\section{Discussion}\label{sec3}

In this large retrospective study, we systematically evaluated the ability of DL-based early warning models to generalise across four distinct ICU data sources. We observed a general drop in model performance when models were applied to a previously unseen data source, sometimes by more than -0.2 in the AUROC. Using multiple data sources during training attenuated the performance gap, providing performance roughly on par with the best model trained on a single data source. We found that applying dedicated algorithms that optimise for generalisability did only haphazardly improve generalisability in our experiments. Futhermore, even if discriminative performance remained stable, models often need to be re-calibrated to the new data source. 
\\[10pt]
Most machine learning or DL-based prediction models to date are not externally validated \cite{shillanUseMachineLearning2019a}. Our results reiterate the importance of external validation for the interpretation of model performance. However, caution is warranted when reporting validation results. Good validation performance may be confounded by differences in case-mix. For example, a sepsis model built on eICU and evaluated on MIMIC may appear stable (AUROC 0.749 vs. 0.769). Yet, achievable performance in MIMIC would have been higher (AUROC 0.824), suggesting that the eICU model is actually suboptimal for MIMIC. If sample size permits, we believe it is thus good practice to also train a model on the external dataset alone and report such "oracle" performance for comparison. 


Our study builds on recent work by Moor \textit{et al.} to harmonise ICU data sources \cite{moorPredictingSepsisMultisite2021, bennetRicuInterfaceIntensive2021}. In a proof-of-concept study, the harmonised data was applied to sepsis prediction where --- in line with our findings --- reduced performance was observed when models were transfered between data sources, particularly when extrapolating from European patients to the US \cite{moorPredictingSepsisMultisite2021}. We showed that similar patterns can be found in other common prediction tasks, alluding to systematic differences. European data may for example cover sicker ICU populations \cite{sauerSystematicReviewComparison2022a}, which may not generalise to less severe patient groups. Moor \textit{et al.} reported that using an ensemble of sepsis models can mitigate the loss in performance \cite{moorPredictingSepsisMultisite2021}. We showed that this is also the case for a single model trained on pooled data. The underlying mechanisms at play may be similar to those during federated learning, where similarly improved generalisability has been reported for imaging tasks \cite{riekeFutureDigitalHealth2020, sarmaFederatedLearningImproves2021}.

The use of dedicated algorithms to promote generalisability of ICU prediction models was first investigated by Zhang \textit{et al.} \cite{zhangEmpiricalFrameworkDomain2021}. They limited their analysis to the eICU dataset --- which itself includes data from 208 hospitals --- and defined domains as US geographical regions (Northeast, Midwest, West, South, Missing). However, their baseline models already generalised well across regions, preventing further assessment of domain generalisation algorithms. A follow-up study by Spathis and Hyland \cite{spathisLookingOutofDistributionEnvironments2022} also struggled to identify non-generalisable domains in eICU. Our results may provide additional context for these unexpected findings. With as few as three training datasets, pooled training was able to considerably improve generalisability in our experiments, matching the performance of the best single-source model. The large number of hospitals in eICU may represent a limiting case of this phenomenon, providing enough variability such that for every hospital in the test data, there is almost always at least one hospital in the training data that is similar. 

Even though we \textit{did} find a generalisation gap in our data, none of the domain generalisation algorithms considered in this study consistently outperformed regular pooled training via ERM. This finding is in keeping with a recent, large benchmark study on image classification \cite{gulrajaniSearchLostDomain2020}. When carefully evaluating state-of-the-art algorithms --- including those used in our study --- Gulrajani et al. did not find any consistent improvement compared to regular training. Where new algorithms have been proposed since and applied to the benchmark, improvements were usually marginal \cite{rameFishrInvariantGradient2021}.

\subsection{Strengths and limitations}

To the best of our knowledge, this is the first study that investigates the generalisability of longitudinal DL across a range of common ICU prediction tasks and data sources. We included all currently available major open-source ICU data sources \cite{sauerSystematicReviewComparison2022a} and used a dedicated preprocessing package to obtain comparable, high-quality data \cite{bennetRicuInterfaceIntensive2021}. All algorithms were evaluated using a peer-reviewed benchmarking framework \cite{gulrajaniSearchLostDomain2020}.

Although the availability of rich, open-source ICU data has improved, there are currently only four such data sources available globally \cite{sauerSystematicReviewComparison2022a}. All data sources were from Western countries and it is unclear whether our findings apply to data from outside the US and Europe. The limited number of data sources may have inhibited the algorithms' ability to detect invariant relationships. Access to more data sources may enable some algorithms to consistently outperform regular training. However, the results from eICU suggest that the generalisability of regular training also improves as the number of available hospitals increases, raising the question whether there exists an optimal number of hospitals at which specialised algorithms confer the most benefit. Generalisability remains an active field of research and new algorithms may be proposed that more consistently outperform regular training. 

We limited model comparison to the AUROC, which we chose for its widespread use and its interpretability across varying levels of prevalence. Other metrics such as the Brier score or the area under the precision-recall curve may lead to different conclusions and could be investigated in future research. Our results do not necessarily reflect clinical applicability. Task labels were retrospectively derived from routine data and may be imprecise due to missing data or measurement errors. There were considerable variations in outcome prevalence across data sources, which may represent genuine differences in patient populations, administrative differences in data quality, or both. Recent results from a prospective evaluation of a sepsis model nevertheless show that models built from such data may improve patient outcomes \cite{adamsProspectiveMultisiteStudy2022}.

\FloatBarrier

\subsection{Conclusion}

Our results suggest that as more and more ICU data sources become publicly available for training, the models derived from them may become increasingly generalisable. Increasing the number of hospitals represented in the training data improves the chance that for every future hospital, at least one training hospital was similar enough to allow for reliable transfer of knowledge. Researchers aiming to develop DL-based ICU prediction models should therefore spend significant time curating a training set that best reflects the anticipated future hospitals at which their model will be deployed.

\section{Methods}\label{sec4}

\subsection{Data sources}

We used retrospective ICU data from four data sources: AUMCdb version 1.0.2 (Netherlands) \cite{thoralSharingICUPatient2021}, HiRID version 1.1.1 (Switzerland) \cite{hylandEarlyPredictionCirculatory2020}, eICU version 2.0 (US) \cite{pollardEICUCollaborativeResearch2018, pollardEICUCollaborativeResearch2019}, and MIMIC IV version 2.0 (US) \cite{johnsonMIMICIV2022}. All data used in this study was recorded as part of routine care. All datasets were collected at a single hospital with the exception of eICU, which is a multi-center dataset of 208 hospitals.  AUMCdb data was accessed through the Data Archive and Networked Services of the Royal Netherlands Academy of Arts and Sciences \cite{paulwgelbersAmsterdamUMCdb}. HiRID, eICU, and MIMIC were accessed through PhysioNet \cite{goldbergerPhysioBankPhysioToolkitPhysioNet2000}. Since the datasets were created independently of each other, they do not share the same data structure or data identifiers. In order to make them interoperable --- and thus usable --- within a common prediction model, we used preprocessing utilities provided by the ricu R package version 0.5.3 \cite{bennetRicuInterfaceIntensive2021}. ricu pre-defines a large number of clinical concepts and how they are load from a given data source, providing a common interface to the data.

\subsection{Study population}

Our study cohort consisted of all patients aged $\geq18$ years that were admitted to the ICU. Each ICU stay was discretised into hourly bins. To guarantee adequate data quality and sufficient data for prediction, we excluded stays with: 1) missing or invalid admission and discharge times, 2) a length of less than six hours in the ICU, 3) recorded measurements in less than four hourly bins and 4) more than 12 consecutive hours in the ICU without a single clinical measurement. The unit of observation were single, continuous ICU stays. As some datasets did not allow to identify which ICU stays belonged to the same patient, all available ICU stays were included in the analysis. Patients could therefore contribute more than one ICU stay. For example, if a patient was discharged to a general ward and subsequently readmitted to the ICU, that patient contributed two ICU stays. 

\subsection{Features and preprocessing}

For each ICU stay, we extracted 52 features (4 static features, 48 time-varying features) recorded during or before admission to the ICU. Static information included age at admission, sex, height, and width. Time-varying features included vital signs and laboratory measurements. No information on comorbidities, medication, or procedures was used in the models. See Supplementary Table \ref{tab:supp-features} for a complete list of features. Features were chosen based on prior literature \cite{meyerMachineLearningRealtime2018, moorPredictingSepsisMultisite2021} and availability across datasets.

Biologically implausible feature values were automatically set to missing by the ricu package \cite{bennetRicuInterfaceIntensive2021}. The remaining raw feature values were screened for systematic differences between data sources. Values were centered and scaled to unit variance. Due to the routine nature of the data, measurements could be missing, for example due to the infrequent measurement of expensive laboratory tests. Missing values were imputed using a last-observation-carried-forward scheme. Missing values before a patient's first recorded value were imputed using the mean value across training samples. To allow the model to distinguish imputed values, missing indicators (0/1) were added to the data. 

\subsection{Tasks}

We considered three prediction tasks: all-cause ICU mortality, onset of AKI, and onset of sepsis. Outcomes and cohorts for these tasks were defined as follows.

\bmhead{ICU mortality} ICU mortality was defined as death while admitted to the ICU. We predicted a patient's risk of dying once per ICU stay using information from the first 24 hours of the stay. To exclude patients that were already dead or about to die at the time of prediction, we excluded all patients who died within 30 hours of ICU admission. We also excluded any patients who were discharged alive during that time for similar reasons. 

\bmhead{Onset of AKI} AKI was defined as KDIGO stage $\geq$1, either due to an increase in serum creatinine or low urine output \cite{kdigoKidneyDiseaseImproving2012}.  At every hour of the ICU stay, we predicted a patient's risk of AKI onset within the next 6 hours using all information available up to that point. We excluded stays in which AKI occurred less than 6 hours after ICU admission. We additionally excluded patients with end-stage renal disease defined as a baseline creatinine $>$4 mg/dL (approximately corresponding to chronic kidney disease stage 5 in a 60-year old male \cite{kdigoKidneyDiseaseImproving2013}).

\bmhead{Onset of sepsis} Following the Sepsis-3 criteria \cite{singerThirdInternationalConsensus2016}, sepsis was defined as organ dysfunction due to infection. Organ dysfunction was operationalised as an increase in Sequential Organ Failure Assessment (SOFA) score of $\geq$2 points. Clinical suspicion of infection was defined as co-occurrence of antibiotic administration and microbiological culture. As little or no microbiology information was available in HiRID and eICU, suspicion of infection in those datasets was defined using only antibiotic administration \cite{moorPredictingSepsisMultisite2021}. At every hour of the ICU stay, we predicted a patient's risk of sepsis onset in the next 6 hours using all information available up to that point.
\\[10pt]
Further details on how outcomes were ascertained from the raw data can be found in the supplement. For both AKI and sepsis prediction, all hours up to discharge, death, or 7 days were included in the prediction, whichever happened first. If disease onset was determined, the previous 6 hours were labelled positive. The following 6 hours were also labelled positive, penalising the model for late false negatives, after which follow-up was stopped. Finally, because data quality among hospitals contributing to eICU has been reported to be variable \cite{moorPredictingSepsisMultisite2021}, we excluded any hospitals with no AKI or sepsis cases (2 [1.0\%] hospitals for AKI and 80 [38.5\%] for sepsis).

\subsection{Deep learning models}

We considered three neural network architectures to process the time series data: gated recurrent units (GRU) \cite{choPropertiesNeuralMachine2014}, temporal convolutional networks (TCN) \cite{baiEmpiricalEvaluationGeneric2018}, and transformers \cite{vaswaniAttentionAllYou2017a}. In each case, the time-sensitive network was followed by a simple feed-forward network to predict the outcome from the embedded time series. We trained all models in pytorch version 1.12.1 using the implementations by Yèche et al. \cite{yecheHiRIDICUBenchmarkComprehensiveMachine2021}. Models were trained on the HPC for Research cluster of the Berlin Institute of Health and the HPC@Charité. 

\subsection{Training and evaluation}
\label{sec:training}

We held out 20\% of the data as test set and used five-fold cross-validation to randomly split the remaining data into training (64\%) and validation (16\%). Ten random draws of hyperparameters were run for each model type and realisation of the dataset (see Table \ref{tab:model-hparams} for a definition of the hyperparameter space for each model). Models were trained for a maximum of 1,000 epochs using a binary cross-entropy loss and an Adam optimiser. We employed early stopping with a patience of ten epochs, retaining the model iteration that had the highest performance on the validation fold. The hyperparameters with the lowest average validation loss across the five folds were chosen. This model was then evaluated on the held-out test data from each dataset using the AUROC. Model calibration was assessed for the model from the first fold using calibration plots. Calibration was assessed after re-calibration using isotonic regression on the validation set \cite{zadroznyTransformingClassifierScores2002}.

\subsection{Domain generalisation}

Domain generalisation refers to the ability of a model to maintain its performance when applied to previously unseen domains \cite{gulrajaniSearchLostDomain2020}. In the case of ICU prediction models, the term domain commonly refers to organisationally distinct units such as a single hospital or data source. Generalisability across hospitals is desirable because it suggests a certain robustness of the model and increases the confidence that it will perform well in future applications.

\bmhead{ERM} Models are commonly trained using a variation of ERM. The training scheme outlined in the previous section is an example of such "regular training". ERM does not explicitly consider generalisability. Instead, it assumes that patients observed during training are similar to those seen during testing. As training and test patients --- by assumption --- come from the same population, minimising the prediction error for the former also minimises error for the latter. 

\bmhead{CORAL} In addition to minimising prediction error, CORAL \cite{sunDeepCORALCorrelation2016a} promotes domain-invariant \textit{representations} by matching internal network activations across domains. Intermediate layers of neural networks may be viewed as feature extraction steps that learn informative features from the raw data. If these features are similar across domains, it may be more likely that they represent stable, meaningful information. To incentivise such a behaviour, CORAL compares covariances of layer activations between domains and penalises the model for differences. 

\bmhead{VREx} VREx \cite{kruegerOutofDistributionGeneralizationRisk2020} follows a similar rationale but aims at domain-invariant \textit{end-to-end predictors} by matching overall model performance rather than intermediate layer activations. The model is penalised if training losses differ substantially between domains, incentivising it to perform comparably across domains. 

\bmhead{Fishr} In a third variation on this theme, Fishr \cite{rameFishrInvariantGradient2021} encourages domain-invariant \textit{learning progress} by matching the covariance of backpropagated gradients across domains. 

\bmhead{MLDG} MLDG pursues a model that can be fine-tuned to any domain with only a few optimisation steps \cite{liLearningGeneralizeMetaLearning2017}. To obtain such a model, it randomly selects some of the training domains as virtual test domains (conceptually similar to the train-test splits in cross-validation). The model is then updated in such a way that an improvement on the training domains also improves its performance on the virtual test domains. 

\bmhead{GroupDRO} GroupDRO focuses on minimising the worst-case loss across training domains \cite{sagawaDistributionallyRobustNeural2019}. This is conceptually similar to the risk extrapolation proposed in VREx \cite{kruegerOutofDistributionGeneralizationRisk2020} but instead of penalising variations in performance relies on putting more weight on domain(s) with the worst performance.
\\[10pt]
In addition to the above strategies, two oracle settings were considered in which the models were given access to data from the test domain during model development. First, models were optimised directly on training data from the test domain. This is what is applied in most single-source studies without external validation. In this case, the testing data is an independent and identically distributed sample of the training distribution and --- given a sufficiently large sample --- represents an upper boundary of achievable model performance on that dataset \cite{gulrajaniSearchLostDomain2020}. Second, a model was optimised using training data from all domains including the test domain. The model again has access to data from the test domain during training, providing an alternative upper boundary. A more in-depth, technical description of each considered strategy can be found in supplement \ref{sec:dg-detail}. Additional hyperparameters introduced by the above strategies are listed in Table \ref{tab:dg-hparams}. For model selection, we used the training-domain validation set as described in Gulrajani \textit{et al.} \cite{gulrajaniSearchLostDomain2020}.

\backmatter

\bmhead{Acknowledgments}

The authors acknowledge the Scientific Computing of the IT Division at the Berlin Institute of Health and at Charité - Universitätsmedizin Berlin for providing computational resources that have contributed to the research results reported in this paper.

\section*{Declarations}

\subsection*{Ethics approval}

All datasets had pre-existing ethical approvals that allowed for retrospective analysis. Access to the datasets was obtained following the respective requirements for each dataset, including training on data governance.

\subsection*{Funding}

This work was supported through a postdoc grant awarded to PR by the Alexander von Humboldt Foundation (Grant Nr. 1221006). This work also received funding from the European Commission via the Horizon 2020 program for PRECISE4Q (No. 777107, lead: DF).


\subsection*{Competing interests}

The authors report no competing interests.

\subsection*{Data availability}

All data used in this study is open-source and can be obtained from the respective data providers following appropriate training and reasonable request \cite{thoralSharingICUPatient2021, hylandEarlyPredictionCirculatory2020, pollardEICUCollaborativeResearch2018, johnsonMIMICIV2022}. 

\subsection*{Code availability}

All analysis code used in this study is available on GitHub. R code used to define the cohorts can be found at \href{https://github.com/prockenschaub/icuDG-preprocessing}. Python code used to run all experiments can be found at \href{https://github.com/prockenschaub/icuDG}.


\bibliography{sn-bibliography}


\appendix
\renewcommand{\thefigure}{S\arabic{figure}}
\setcounter{figure}{0}
\renewcommand{\thetable}{S\arabic{table}}
\setcounter{table}{0}
\newpage 

\section{Exclusion criteria}

\begin{figure}[ht]
\centering
\includegraphics[width=\textwidth]{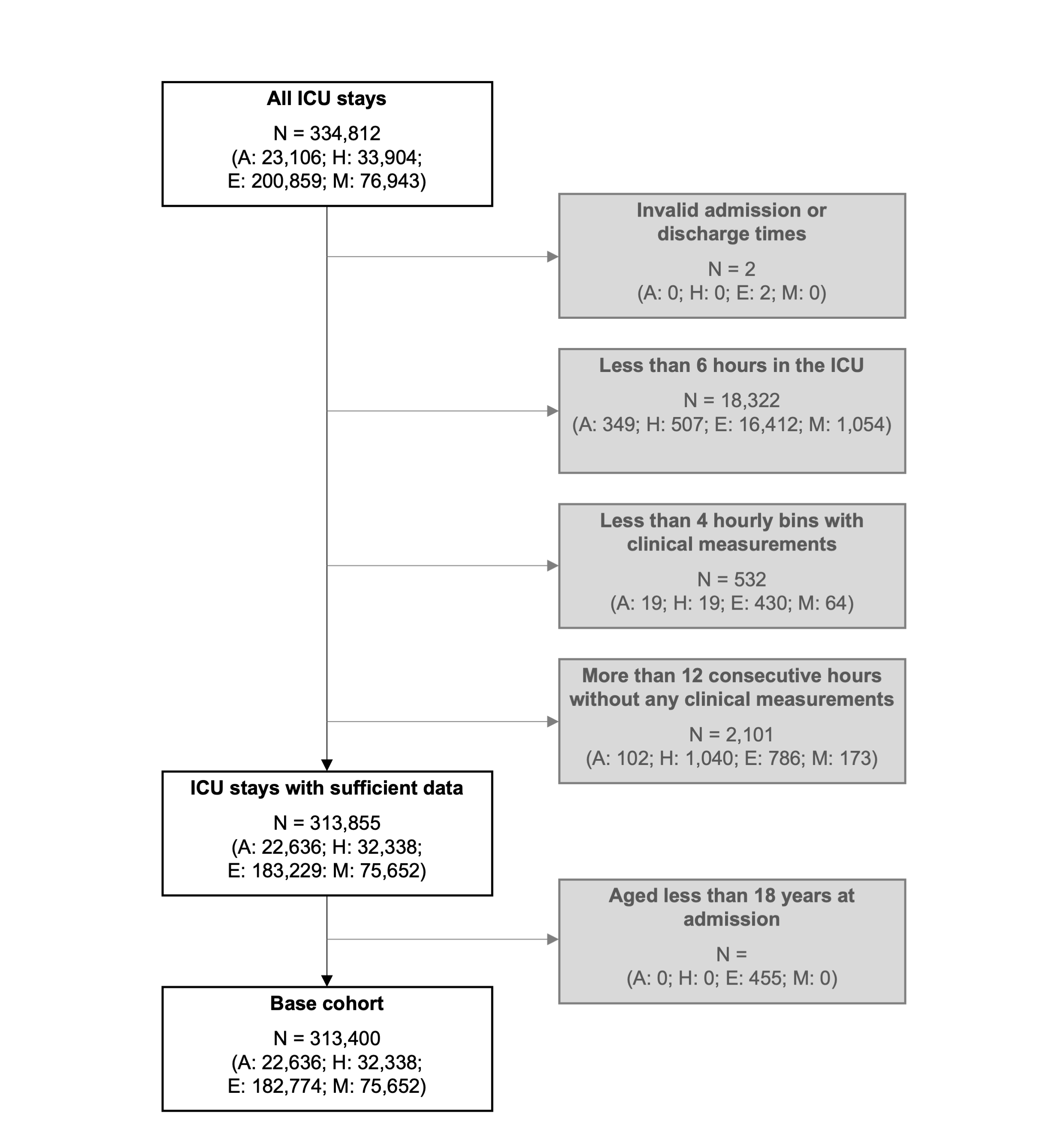}
\caption[short]{Exclusion criteria applied to the base cohort.}
\label{fig:attrition-base}
\end{figure}

We included all available ICU stays of adult patients in our analysis. For each stay, we applied the following exclusion criteria to ensure sufficient data volume and quality: remove any stays with 1) an invalid admission or discharge time defined as a missing value or negative calculated length of stay, 2) less than six hours spent in the ICU, 3) less than four separate hours across the entire stay where at least one feature was measured, 4) any time span of $\geq$12 consecutive hours throughout the stay during which no feature was measured. Figure \ref{fig:attrition-base} details the number of stays overall and by dataset that were excluded this way.

\begin{sidewaysfigure}
\centering
\includegraphics[width=0.9\textwidth]{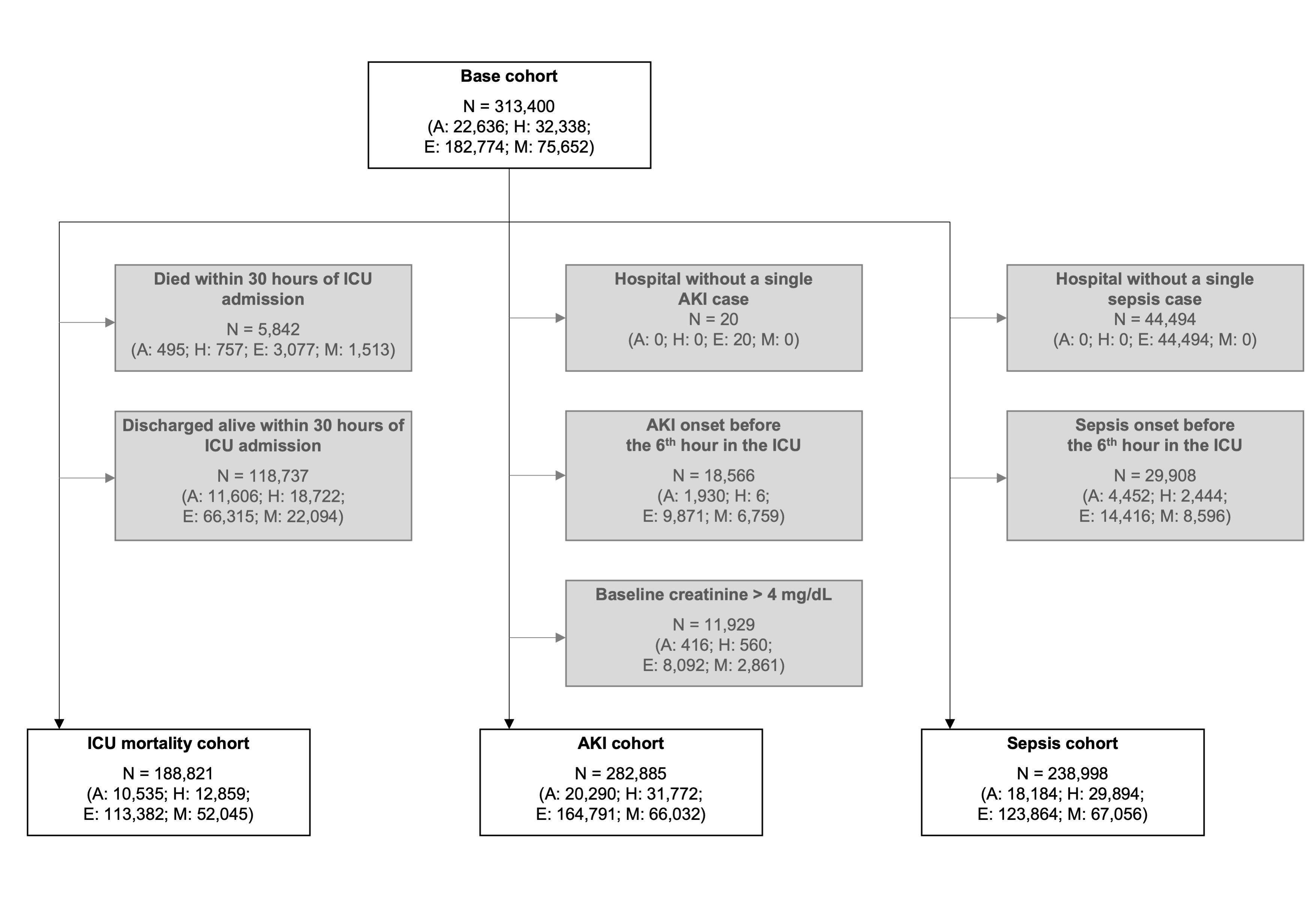}
\caption[short]{Additional exclusion criteria applied depending on the task.}
\label{fig:attrition-tasks}
\end{sidewaysfigure}

Additional exclusion criteria were applied based on the individual tasks (Figure \ref{fig:attrition-tasks}). For ICU mortality, we excluded all patients with a length of stay of less than 30 hours (either due to death or discharge). A minimum length of 30 hours was chosen in order to exclude any patients that were about to die (the sickest patients) or be discharged (the healthiest patients) at the time of prediction at 24 hours. For AKI and sepsis, we excluded any stays were disease onset was outside of ICU or within the first six hours of the ICU stay. To account for differences in data recording across hospitals in eICU, we further excluded hospitals that did not have a single patient with AKI or sepsis to exclude hospitals with insufficient recording of features necessary to define the outcome. Finally for the AKI task, we excluded stays were the baseline creatinine, defined as the last creatinine measurement prior to ICU (if exists) or the earliest measurement in the ICU, was $\gt$4mg/dL to exclude patients with preexisting renal insufficiency.

\section{Feature definitions and units}

A total of 52 features were used for model training (Table \ref{tab:supp-features}), 4 of which were static and 48 that were dynamic. Dynamic features included vital signs (7 variables) and laboratory tests (39 variables), with two more variables that measure input (fraction of inspired oxygen) and output (urine). All variables were extracted via the \href{ricu}{https://cran.r-project.org/web/packages/ricu/index.html} R package. The ricu concept name for each feature is shown in Table \ref{tab:supp-features}. The exact definition of each feature and how it was extracted from the individual databases can be found in the concept configuration file of ricu's GitHub repository (commit \href{885bd0c}{https://github.com/eth-mds/ricu/blob/09902bdc57a1f2f720a924d32f5e053ce2ce7f97/inst/extdata/config/concept-dict.json}).

\begin{longtable}{p{0.58\textwidth}p{0.10\textwidth}p{0.22\textwidth}}
    \caption{Clinical concepts used as input to the prediction models}\label{tab:supp-features}\\
    \hline
     Feature & ricu & Unit  \\
    \hline
    \endfirsthead
    
    \caption{Clinical concepts used as input to the prediction models (continued)}\\
    \hline
     Feature & ricu & unit  \\
    \hline
    \endhead
    
    \textbf{Static} &  &  \\
    Age at hospital admission & age & Years \\
    Female sex & sex & - \\
    Patient height & height & cm \\
    Patient weight & weight & kg \\
    & & \\
    \textbf{Time-varying} &  &  \\
    Blood pressure (systolic) & sbp & mmHg \\
    Blood pressure (diastolic) & dbp & mmHg \\
    Heart rate & hr & beats/minute \\
    Mean arterial pressure & map & mmHg \\
    Oxygen saturation & o2sat & \% \\
    Respiratory rate & resp & breaths/minute \\
    Temperature & temp & $^{\circ}$C \\
    & & \\
    Albumin & alb & g/dL \\
    Alkaline phosphatase & alp & IU/L \\
    Alanine aminotransferase & alt & IU/L \\
    Aspartate aminotransferase & ast & IU/L \\
    Base excess & be & mmol/L \\
    Bicarbonate & bicar & mmol/L \\
    Bilirubin (total) & bili & mg/dL \\
    Bilirubin (direct) & bili\_dir & mg/dL \\
    Band form neutrophils & bnd & \% \\
    Blood urea nitrogen & bun & mg/dL \\
    Calcium & ca & mg/dL \\
    Calcium ionized & cai & mmol/L \\
    Creatinine & crea & mg/dL \\
    Creatinine kinase & ck & IU/L \\
    Creatinine kinase MB & ckmb & ng/mL \\
    Chloride & cl & mmol/L \\
    CO$^2$ partial pressure & pco2 & mmHg \\
    C-reactive protein & crp & mg/L \\
    Fibrinogen & fgn & mg/dL \\
    Glucose & glu & mg/dL \\
    Haemoglobin & hgb & g/dL \\
    International normalised ratio (INR) & inr\_pt & - \\
    Lactate & lact & mmol/L \\
    Lymphocytes & lymph & \% \\
    Mean cell haemoglobin & mch & pg \\
    Mean corpuscular haemoglobin concentration & mchc & \% \\
    Mean corpuscular volume & mcv & fL \\
    Methaemoglobin & methb & \% \\
    Magnesium & mg & mg/dL \\
    Neutrophils & neut & \% \\
    O$^2$ partial pressure & po2 & mmHg \\
    Partial thromboplastin time & ptt & sec \\
    pH of blood & ph & - \\
    Phosphate & phos & mg/dL \\
    Platelets & plt & 1,000 / $\mu$L \\
    Potassium & k & mmol/L \\
    Sodium & na & mmol/L \\
    Troponin T & tnt & ng/mL \\
    White blood cells & wbc & 1,000 / $\mu$L \\
    & & \\
    Fraction of inspired oxygen & fio2 & \% \\
    Urine output & urine & mL \\
    \hline
\end{longtable}

\section{Outcome definitions}
\label{sec:outc-detail}

\subsection{ICU mortality}
ICU mortality was defined as death while in the ICU. This was generally ascertained via the recorded discharge status or discharge destination. Note that our definition of ICU mortality differs from the definition of \verb|death| in the ricu R package, which describes hospital mortality that is not available for all included datasets.

\bmhead{AUMCdb} Death was inferred from the \verb|destination| column of the \verb|admissions| table. A destination of "Overleden" (Dutch for "passed away") was treated as death in the ICU. Since the date of death was recorded outside of the ICU and may therefore be imprecise, the recorded ICU discharge date was used as a more precise proxy for the time of death. 

\bmhead{HiRID} Death was inferred from the column \verb|discharge_status| in table \verb|general|. A status of "dead" was treated as death in the ICU. Time of death was inferred as the last measurement of IDs \verb|110| (mean arterial blood pressure) or \verb|200| (heart rate) in column \verb|variableid| of table \verb|observations|.

\bmhead{eICU} Death was inferred from the column \verb|unitdischargestatus| in table \verb|patient|. A status of "Expired" was treated as death in the ICU. The recorded ICU discharge date was used as a proxy for the time of death.

\bmhead{MIMIC IV} Death was inferred from the column \verb|hospital_expire_flag| in table \verb|admissions|. Since MIMIC IV only records a joint ICU/hospital expire flag, ward transfers were analysed to ascertained location of death. If the last ward was an ICU ward, the death was considered ICU mortality.

\subsection{Acute kidney injury}
AKI was defined as KDIGO stage $\geq$1, either due to an increase in serum creatinine or low urine output (Table \ref{tab:kdigo}) \cite{kdigoKidneyDiseaseImproving2012}. Baseline creatinine was defined as the lowest creatinine measurement over the last 7 days. Urine rate was calculated as the amount of urine output in ml divided by the number of hours since the last urine output measurement (for a max gap of 24h), except for HiRID in which urine rate was recorded directly. The earliest urine output was divided by 1. Rate per kg was calculated based on the admission weight. If weight was missing, a weight of 75kg was assumed instead. 

\bmhead{AUMCdb} Creatinine was defined via the standard ricu concept of serum creatinine as IDs \verb|6836|, \verb|9941|, or \verb|14216| in column \verb|itemid| of table \verb|numericitems|. Urine output was defined as IDs \verb|8794|, \verb|8796|, \verb|8798|, \verb|8800|, \verb|8803| in column \verb|itemid| of table \verb|numericitems| (note that this includes more items than those included in the standard ricu concept of urine output). 

\bmhead{HiRID} Creatinine was defined via the standard ricu concept of serum creatinine as ID \verb|20000600| in column \verb|variableid| of table \verb|observations|. Urine rate was defined as ID \verb|10020000| in column \verb|variableid| of table \verb|observations|. 

\bmhead{eICU} Creatinine was defined via the standard ricu concept of serum creatinine as IDs "creatinine" in column \verb|labname| of table \verb|lab|. Urine output was defined via the standard ricu concept of urine output as IDs "Urine" and "URINE CATHETER" in column \verb|celllabel| of table \verb|intakeoutput|.

\bmhead{MIMIC IV} Creatinine was defined via the standard ricu concept of serum creatinine as ID \verb|50912| in column \verb|itemid| of table \verb|labevents|. Urine output was defined via the standard ricu concept of urine output as IDs \verb|226557|, \verb|226558|, \verb|226559|, \verb|226560|, \verb|226561|, \verb| 226563|, \verb|226564|, \verb|226565|, \verb|226566|, \verb|226567|, \verb|226584|, \verb|227510| in column \verb|itemid| of table \verb|outputevents|.

\begin{table}[ht]
    \scriptsize
    \caption{Staging of AKI according to KDIGO \cite{kdigoKidneyDiseaseImproving2012}}
    \label{tab:kdigo}
    \begin{threeparttable}
        \begin{tabular}{llcc}
            \toprule
            \textbf{Stage} & \textbf{Serum creatinine} & \textbf{Urine output}  \\
            \midrule
            1 & \makecell{1.5–1.9 times baseline \\\\ OR \\\\ $\geq$0.3 mg/dl ($\geq$26.5 $\mu$mol/l) increase \\ within 48 hours} & $\lt$0.5 ml/kg/h for 6–12 hours \\
            \midrule
            2 & \makecell{2.0–2.9 times baseline } & $\lt$0.5 ml/kg/h for $\geq$12 hours \\
            \midrule
            3 & \makecell{3.0 times baseline (prior 7 days) \\\\ OR \\\\ Increase in serum creatinine to \\ $\geq$4.0 mg/dl ($\geq$353.6 $\mu$mol/l) \\ within 48 hours \\\\ OR \\\\ Initiation of renal replacement therapy} & \makecell{$\lt$0.3 ml/kg/h for $\geq$24 hours \\\\ OR \\\\ Anuria for $\geq$12 hours}\\
            \bottomrule
        \end{tabular}
        \begin{tablenotes}[flushleft, para]
            \footnotesize
            \item \leavevmode\kern-\scriptspace\kern-\labelsep AKI, acute kidney injury; KDIGO, Kidney Disease Improving Global Outcomes.
        \end{tablenotes}
    \end{threeparttable}
\end{table}

\subsection{Sepsis}

Onset of sepsis was defined using the Sepsis-3 criteria \cite{singerThirdInternationalConsensus2016}, which defines sepsis as organ dysfunction due to infection. Following guidance from the original authors of Sepsis-3 \cite{seymourAssessmentClinicalCriteria2016}, organ dysfunction was defined as an increase in SOFA score $\geq$2 points compared to the lowest value over the last 24 hours. Suspicion of infection was defined as a simultaneous use of antibiotics and culture of body fluids. The time of sepsis onset was defined as the first time of organ dysfunction within 48 hours before and 24 hours after suspicion of infection. Time of suspicion was defined as the earlier of antibiotic initiation or culture request. Antibiotics and culture were considered concomitant if the culture was requested $\leq$24 hours after antibiotic initiation or if antibiotics were started $\leq$72 hours after the culture was sent to the lab. Where available, antibiotic treatment was inferred from administration records, otherwise we used prescription data. In order to exclude prophylactic antibiotics, we required that antibiotics were administered continuously for $\geq$3 days \cite{reynaEarlyPredictionSepsis2020}. Antibiotics treatment was considered continuous if an antibiotic was administered once every 24 hours for 3 days (or until death) or were prescribed for the entire time spent in the ICU. HiRID and eICU did not contain microbiological information. For these datasets, we followed Moor et al.  \cite{moorPredictingSepsisMultisite2021} and defined suspicion of infection through antibiotics alone. Note, however, that the sepsis prevalence in our study was considerably lower than theirs, which was as high as 37\% in HiRID. We suspect that this is because they did not require treatment for $\geq$3 days.

\bmhead{AUMCdb} The SOFA score, microbiological cultures, and antibiotic treatment were defined via the standard ricu concepts \verb|sofa|, \verb|abx|, and \verb|samp| (see the \href{ricu}{https://github.com/eth-mds/ricu/blob/09902bdc57a1f2f720a924d32f5e053ce2ce7f97/inst/extdata/config/concept-dict.json} package for more details). 

\bmhead{HiRID} The SOFA score and antibiotic treatment were defined via the standard ricu concept \verb|sofa| and \verb|abx| (see the \href{ricu}{https://github.com/eth-mds/ricu/blob/09902bdc57a1f2f720a924d32f5e053ce2ce7f97/inst/extdata/config/concept-dict.json} package for more details). No microbiology data was available in HiRID. 

\bmhead{eICU} The SOFA score and antibiotic treatment were defined via the standard ricu concept \verb|sofa| and \verb|abx| (see the \href{ricu }{https://github.com/eth-mds/ricu/blob/09902bdc57a1f2f720a924d32f5e053ce2ce7f97/inst/extdata/config/concept-dict.json} package for more details). Microbiology data in eICU was not reliable \cite{moorPredictingSepsisMultisite2021} and therefore omitted.

\bmhead{MIMIC IV} The SOFA score and microbiological cultures were defined via the standard ricu concepts \verb|sofa| and \verb|samp| (see the \href{ricu}{https://github.com/eth-mds/ricu/blob/09902bdc57a1f2f720a924d32f5e053ce2ce7f97/inst/extdata/config/concept-dict.json} package for more details). Antibiotics were defined based on table \verb|inputevents|. This differs from the standard ricu \verb|abx| concept, which also considers the \verb|prescriptions| table.

\section{Domain generalisation methods}
\label{sec:dg-detail}

Consider a dataset $D = \{x_i, y_i\}^n_{i=1}$ containing data on $n$ patients, where $x_i \in \mathbb{R}^{T_i \times P}$ are the observed $P$ feature values for patient $i$ at each time step $t = \{1, ..., T_i\}$ and $y_i$ is the outcome of interest. If we are only interested in making a single prediction at a given time point (as is the case in mortality prediction after 24 hours), $y_i \in \{0, 1\}$ is a scalar value indicating the presence or absence of the outcome. Alternatively, we may want to make a prediction at every time step (as is the case for predicting onset of AKI or sepsis), in which case $y_i \in \{0, 1\}^{T_i}$. Our objective is to train a model $f_{\theta}$ that uses each patient's features $x_i$ to accurately predict $y_i$.

In most conventional settings, we assume that our sample $D$ was drawn from common distribution $P(x, y)$. We make the same assumption about any future samples $D'$. As we argue in the introduction, this assumption is often violated in practice. If we collect additional data $D' \sim P_{D'}$ from a new hospital, it is likely that it's distribution will differ in some aspects, i.e., $P_{D'}(x, y) \neq P(x, y)$. Instead of a single data-generating distribution $P(x,y)$, we may therefore think of data $D_e$ collected at hospital $e$ coming from domain-specific distributions $D_e \sim P_e(x, y)$, with $P_e(x, y) \in \{P_1, P_2, ...\}$ drawn from a large set of possible distributions. Since we can't know in advance what $P_e$ of future hospitals will look like, our objective moves from optimising $f_{\theta}$ for a single $P(x, y)$ to optimising it for any distribution in the set of possible distributions $\{P_1, P_2, ...\}$. In the following, we introduce several methods that have been proposed to solve this problem. 

Throughout this section, we will assume that we are training a neural network predictor that consists of two parts: 1) a featurizer network $f_{\phi}$ that takes the raw input $\{ x_i\}$ and transforms it into an internal representation $\{ z_i\} = \{f_{\phi}( x_i)\}$ and 2) a classifier network $f_w$ that uses the internal representation $ z_i$ to predict the binary outcome $y_i$. For notational simplicity, we will denote the full set of network parameters as $\theta = \{\phi, w\}$ and the end-to-end prediction model as $f_{\theta}(x) = f_w(f_{\phi}(x))$. In this study, the featurizer $f_{\phi}$ was modelled using a time-aware network architecture whereas the classifier $f_w$ was a simple feed-forward network. The prediction loss was a binary cross-entropy loss.  

\subsection{Empirical Risk Minimisation (ERM)}

ERM minimises the average empirical loss (=risk) across all training domains \cite{vapnikOverviewStatisticalLearning1999}. Following the notation of Rame et al. \cite{rameFishrInvariantGradient2021}, we have 

$$
\mathcal{L}_{ERM}(\theta) = \frac{1}{\lvert \mathcal{E} \rvert }\sum_{e \in \mathcal{E}}\mathcal{L}_e(\theta) 
$$

\noindent where $\mathcal{E} = \{1, 2, ..., \lvert \mathcal{E} \rvert \}$ is the set of training domains and $\mathcal{L}_e(\theta)$ is the empirical loss in training domain $e$ given network parameters $\theta$:

$$
\mathcal{L}_e(\theta) = \frac{1}{n_e} \sum^{n_e}_{i=1} \ell \left( f_{\theta}( x_e^i),  y_e^i \right)
$$

\noindent Each domains empirical loss $\mathcal{L}_e$ is the mean of the prediction losses $\ell$ of all samples $\{ x_e^i,  y_e^i\}$ in that domain. ERM is therefore equivalent to standard pooled training (i.e., when we combine all domains to a single dataset and ignore domain origin) with equal sample sizes $n = n_e$ in each domain or weighted training where each sample is weighted by the inverse of its relative domain size $n_e/n$. 

\subsection{Correlation Alignment (CORAL)}

CORAL augments the ERM loss by adding a second loss term that acts as a regulariser and increases with the magnitude of differences of layer activation distributions between domains. 

$$
\mathcal{L}_{CORAL}(\theta, \gamma) = \mathcal{L}_{ERM}(\theta) + \gamma \mathcal{L}_{augm}(\theta)
$$

\noindent where $\gamma$ is a hyperparameter that governs the strength of the regularisation. Sun and Saenko \citep{sunDeepCORALCorrelation2016a} defined $\mathcal{L}_{augm}(\theta)$ as the distance between the second-order statistics (=covariances) of a pair of domains $A$ and $B$

$$
\mathcal{L}_{augm}(\theta) = \frac{1}{4d^2}\|Cov_A(\theta)-Cov_B(\theta)\|_F
$$

\noindent where $\|\cdot\|_F$ denotes the squared matrix Frobenius norm. The covariance $C_e$ of a domain $e$ is given by:

$$
Cov_e(\theta) = \frac{1}{n-1} \left( z_e^\intercal z_e - \frac{1}{n}(\mathbbm{1}^\intercal z_e)^\intercal (\mathbbm{1}^\intercal z_e) \right)
$$

\noindent where $\mathbbm{1}$ is a column vector with all elements equal to $1$ and $z_e \in \mathcal{R}^{n\times d} $ is the matrix of intermediate layer activations after applying the featuriser network $f_\phi$ for a batch of size $n$ .

The CORAL implementation in the DomainBed framework \cite{gulrajaniSearchLostDomain2020} (and therefore the one employed in this study) uses a variant of the loss proposed by Sun and Saenko \citep{sunDeepCORALCorrelation2016a}, removing the scaling factor $1/4d^2$ and adding the difference in mean activations $\mu_*$. Since we deal with more than one training domain -- the original paper only considered domain adaption in the case of a source and target domain \citep{sunDeepCORALCorrelation2016a} -- we also need to calculate the difference for each pair $A, B \in \mathcal{E}$, giving the following loss augmentation term:

$$
\mathcal{L}^*_{augm}(\theta) = \sum_{A, B \in \mathcal{E}} \| \mu_A(\theta) - \mu_B(\theta) \| + \|Cov_A(\theta)-Cov_B(\theta)\|_F
$$

\noindent In the case of sequence-to-sequence classifications considered in our study, the internal representation $z_e \in \mathcal{R}^{n \times t \times d} $ is 3-dimensional with the added dimension of time $t$. We extended CORAL to this setting by treating each patient's time step as independent observations, reducing the problem to standard classification. We did so by reshaping the representation to $\mathcal{R}^{nt \times d}$ and using the existing equations outlined above.  

\subsection{Variance Risk Extrapolation (VREx)}

VREx adds a different loss augmentation term to ERM that -- rather than encouraging invariant network-internal representation like CORAL -- aims to create a model whose performance is invariant across domains \cite{kruegerOutofDistributionGeneralizationRisk2020}. Differences in model performance are quantified via the variance in the empirical loss $\mathcal{L}_e$ across domains.

$$
\mathcal{L}_{augm}(\theta) = Var \left( \{ \mathcal{L}_1(\theta), \mathcal{L}_2(\theta), ..., \mathcal{L}_{\lvert \mathcal{E} \rvert }(\theta)\} \right)
$$

\noindent As with CORAL, the total loss function is a weighted sum of ERM loss and the augmentation term. 

$$
\mathcal{L}_{VREx}(\theta, \lambda) = \mathcal{L}_{ERM}(\theta) + \lambda \mathcal{L}_{augm}(\theta)
$$

\noindent Note that the weight for VREx is called $\beta$ in the original paper \cite{kruegerOutofDistributionGeneralizationRisk2020} and $\lambda$ in DomainBed \cite{gulrajaniSearchLostDomain2020}. The notation here follows that of DomainBed. VREx additionally allows for a warm-up period during which $\lambda=0$. This allows the classifier to learn a preliminary model using only the ERM loss before it is penalised. The number of warm-up steps is a hyperparameter and can be tuned. 

\subsection{Fishr}

Fishr augments ERM with a loss term that depends on the distribution of gradients across domains \cite{rameFishrInvariantGradient2021}. The networks' gradients are the derivative of the loss function with respect to the network weights and measure how the loss would change if we changed the value of a weight by a very small amount (holding all others constant). Gradients are usually used during back-propagation to update model parameters. Comparable gradients across domains may thus be seen as evidence of a shared mechanism being learnt \cite{rameFishrInvariantGradient2021}. Differences in gradients are measured by the gradient variances

$$
\mathcal{L}_{augm}(\theta) = \frac{1}{\lvert \mathcal{E} \rvert} \sum_{e \in \mathcal{E}} \|  v_e - {\bar v}\|
$$

\noindent where $\|\cdot\|$ is the Euclidean distance, $ v_e = \frac{1}{n_e -1} \sum_{i=1}^{n_e} ( g^i_e - {\bar g_e})^2$ is the gradient variance, and $ g^i_e = \nabla_w \ell \left( f_w( x^i_e)  y^i_e \right)$ is the gradient vector for sample $i$. Note that in order to be computationally feasible, the gradients are only calculated for the classifier network $f_w$. Furthermore, although Rame et al. \cite{rameFishrInvariantGradient2021} theoretically motivate the comparison of the domains' Hessian matrices, this is only approximated through gradient variances to reduce computational complexity. Like with CORAL and VREx, the final loss function is a weighted sum of the ERM loss and the augmentation term.

$$
\mathcal{L}_{Fishr}(\theta, \lambda) = \mathcal{L}_{ERM}(\theta) + \lambda \mathcal{L}_{augm}(\theta)
$$

The algorithm allows for a warm-up period during which $\lambda=0$. Rame et al. \cite{rameFishrInvariantGradient2021} further proposed smoothing the variance at each update step $t$ with a moving average ${\bar v^t_e} = \gamma {\bar v^{t-1}_e} + (1 - \gamma){v^{t}_e}$.

\subsection{Meta-Learning for Domain Generalisation (MLDG)}

MLDG takes a different approach and casts domain generalisation into a meta-learning framework \cite{liLearningGeneralizeMetaLearning2017}. It aims to find a model that "learns to learn". To do so, at each learning iteration we split the training domains $\mathcal{E}$ into a set of meta-test domains $\mathcal{V}$ and a set of meta-train domains $\mathcal{S} = \mathcal{E} - \mathcal{V}$. We proceed by calculating the standard ERM loss for the meta-train domains.

$$
\mathcal{L}_{\mathcal{S}}(\theta) = \frac{1}{\lvert \mathcal{S} \rvert }\sum_{e \in \mathcal{S}}\mathcal{L}_e(\theta) 
$$

\noindent We then create an intermediate network $f_{\theta'}$ by updating the $\theta$ with respect to this loss. For simplicity and clarity, we will use stochastic gradient descent (SGD) for this example, in which case the updated parameters $\theta'$ are obtained as $\theta' = \theta - \alpha \nabla_{\theta}$, where $\alpha$ is the learning rate and $\nabla_{\theta}$ are the gradients of $\theta$ with respect to $\mathcal{L}_{\mathcal{S}}(\theta)$. The updated model is evaluated on the meta-test domains as

$$
\mathcal{L}_{\mathcal{V}}(\theta') = \frac{1}{\lvert \mathcal{V} \rvert }\sum_{e \in \mathcal{V}}\mathcal{L}_e(\theta') 
$$

\noindent This evaluation loss measures how well the intermediate network $f_{\theta'}$ --- which was refined for the meta-train domains --- works on the meta-test domains. If the update only improve the former but degraded performance on the meta-test domains, this would be reflected in the gradients of this loss. The actual model optimisation is therefore taken with respect to both losses 

$$
\mathcal{L}_{MLDG}(\theta, \beta) = \mathcal{L}_{\mathcal{S}}(\theta) + \beta \mathcal{L}_{\mathcal{V}}(\theta')
$$

\noindent where $\beta$ is a weight term that controls the relative importance of the evaluation loss. Sticking with SGD, the updated parameter $\theta^*$ after the iteration is calculated as $\theta^* = \theta - \alpha \nabla_{\theta} \mathcal{L}_{MLDG}(\theta, \beta)$.  Note that this requires calculating a gradient through a gradient because $\nabla_{\theta}\mathcal{L}_{\mathcal{V}}(\theta') = \nabla_{\theta}\mathcal{L}_{\mathcal{V}}(\theta - \alpha \nabla_{\theta}\mathcal{L}_{\mathcal{S}}(\theta))$. As this is computationally expensive, the DomainBed \cite{gulrajaniSearchLostDomain2020} implementation uses the following first order approximation: 

$$
\theta^* = \theta - \alpha \left( \nabla_{\theta} \mathcal{L}_{\mathcal{S}}(\theta) + \beta \nabla_{\theta'} \mathcal{L}_{\mathcal{V}}(\theta') \right)
$$

\subsection{Group Distributionally Robust Optimisation (GroupDRO)}

GroupDRO tries to minimise the worst-case loss on the training domains \cite{sagawaDistributionallyRobustNeural2019}. This is achieved by dynamically weighting the contributions of each training domain proportionally to their loss. 

$$
\mathcal{L}_{GroupDRO}(\theta, q) = \sum_{e \in \mathcal{E}} q_e \mathcal{L}_{ERM}(\theta)
$$

where $q \in \Delta_{\lvert \mathcal{E} \rvert}$ is a $(\lvert \mathcal{E} \rvert-1)$-dimensional probability simplex that governs the relative contribution of each domain to the loss. $q$ is learned together with $\theta$ using exponential gradient ascent $q_e' = q_e e^{\eta \mathcal{L}_e}$ followed by renormalisation $q_e' = q_e' / \sum_{e \in \mathcal{E}} q_e'$. The learning rate for $q$ is determined by $\eta$, which is a hyperparameter.

\section{Hyperparameters and additional results}

\begin{table}[ht]
    \scriptsize
    \caption{Model hyperparameters, their default values, and distributions considered during random search}
    \label{tab:model-hparams}
    \begin{tabular}{llcc}
        \toprule
         & \textbf{Parameter} & \textbf{Default} & \textbf{Random distribution} \\
        \midrule
        \multirow{4}{*}{All models} & Learning rate & 0.001 & e\textsuperscript{Uniform(-10, -3)} \\
         & Weight decay & 0 & Choice([0, 10e-7, 10e-6, ..., 10e0]) \\
         & Dropout probability & 0.5 & Choice([0.3, 0.4, 0.5, 0.6, 0.7]) \\
         & Batch size & 128 & Choice([128, 256, 512])\\
        \midrule
        \multirow{2}{*}{GRU} & Hidden dimension & 64 & Choice([32, 64, 128]) \\
        & Number of layers & 1 & RandomInt(1, 10) \\
        \midrule
        \multirow{3}{*}{TCN} & Hidden dimension & 64 & Choice([32, 64, 128]) \\
        & Number of layers & 1 & RandomInt(1, 10) \\
        & Kernel size & 4 & RandomInt(2, 6) \\
        \midrule
        \multirow{3}{*}{Transformer} & Hidden dimension & 64 & Choice([32, 64, 128]) \\
        & Number of layers & 1 & RandomInt(1, 10) \\
        & Number of heads & 4 & RandomInt(1, 6) \\
        \bottomrule
    \end{tabular}
\end{table}

\begin{table}[ht]
    \scriptsize
    \caption{Domain generalisation hyperparameters, their default values, and distributions considered during random search}
    \label{tab:dg-hparams}
    \begin{tabular}{llcc}
        \toprule
         & \textbf{Parameter} & \textbf{Default} & \textbf{Random distribution} \\
        \midrule
        CORAL & Loss weight $\gamma$ & 1,000 & 10\textsuperscript{Uniform(2, 4)} \\
        \midrule
        \multirow{2}{*}{VREx} & Loss weight $\lambda$ & 1,000 & 10\textsuperscript{Uniform(2, 4)} \\
        & Number of warm-up steps & 100 & 10\textsuperscript{Uniform(0, 3)} \\
        \midrule
        \multirow{2}{*}{Fishr} & Loss weight $\lambda$ & 1,000 & 10\textsuperscript{Uniform(2, 4)} \\
        & Number of warm-up steps & 100 & 10\textsuperscript{Uniform(0, 3)} \\
        \midrule
        \multirow{2}{*}{MLDG} & Loss weight $\beta$ & 1 & 10\textsuperscript{Uniform(-1, 1)} \\
        & Number of virtual test envs & 2 & Choice([1, 2]) \\
        \midrule
        GroupDRO & Exponential update rate $\eta$ & 0.01 & 10\textsuperscript{Uniform(-3, -1)} \\
        \bottomrule
    \end{tabular}
\end{table}

\begin{table}[ht]
    \scriptsize
    \caption{Mean AUROC and standard errors achieved by different network architectures when trained to predict ICU mortality on one dataset (rows) and evaluated on others (columns). }
    \label{tab:erm-mort-detail}
    \centering
    \begin{threeparttable}
        \begin{tabularx}{\textwidth}{Xcccr}
            \toprule
            \textbf{Training domain} & \multicolumn{4}{c}{\textbf{Test domain}} \\\cline{2-5}
             & AUMCdb & HiRID & eICU & MIMIC \\
            \midrule
            \addlinespace[0.3em]
            \multicolumn{5}{l}{\textbf{Gated Recurrent Network}}\\
            AUMCdb & 0.839±0.007 & 0.746±0.008 & 0.776±0.009 & 0.787±0.005 \\ 
            HiRID & 0.753±0.011 & 0.859±0.007 & 0.730±0.010 & 0.756±0.006 \\ 
            eICU & 0.825±0.008 & 0.800±0.005 & 0.856±0.003 & 0.848±0.003 \\ 
            MIMIC & 0.816±0.004 & 0.808±0.006 & 0.821±0.001 & 0.869±0.002 \\ 
            pooled (n-1) & 0.827±0.006 & 0.785±0.007 & 0.828±0.002 & 0.839±0.002 \\ 
            all & 0.842±0.006 & 0.864±0.004 & 0.850±0.002 & 0.869±0.004 \\ 
            \midrule
            \multicolumn{5}{l}{\textbf{Temporal Convolutional Network}}\\
            AUMCdb & 0.848±0.006 & 0.781±0.006 & 0.786±0.006 & 0.792±0.005 \\ 
            HiRID & 0.751±0.015 & 0.839±0.008 & 0.681±0.035 & 0.710±0.021 \\ 
            eICU & 0.815±0.004 & 0.796±0.009 & 0.853±0.001 & 0.846±0.003 \\ 
            MIMIC & 0.810±0.006 & 0.808±0.008 & 0.828±0.003 & 0.873±0.003 \\ 
            pooled (n-1) & 0.834±0.004 & 0.800±0.008 & 0.827±0.001 & 0.837±0.002 \\ 
            all & 0.842±0.007 & 0.863±0.005 & 0.848±0.003 & 0.862±0.003 \\ 
            \midrule
            \multicolumn{5}{l}{\textbf{Transformer}}\\
            AUMCdb & 0.816±0.007 & 0.717±0.011 & 0.734±0.022 & 0.736±0.022 \\ 
            HiRID & 0.715±0.030 & 0.844±0.002 & 0.698±0.023 & 0.713±0.022 \\ 
            eICU & 0.799±0.006 & 0.774±0.007 & 0.845±0.002 & 0.826±0.007 \\ 
            MIMIC & 0.796±0.006 & 0.763±0.008 & 0.807±0.002 & 0.854±0.003 \\ 
            pooled (n-1) & 0.828±0.006 & 0.775±0.008 & 0.824±0.001 & 0.827±0.006 \\ 
            all & 0.849±0.001 & 0.858±0.004 & 0.846±0.005 & 0.855±0.004 \\ 
            \bottomrule
        \end{tabularx}
        \begin{tablenotes}[flushleft, para]
            \footnotesize
            \item \leavevmode\kern-\scriptspace\kern-\labelsep Standard errors were calculated based on 5-fold cross-validation as $se_{cv} = \sigma_{cv}n^{-1/2}$, where $\sigma_{cv}$ is the standard deviation across cross-validation folds and $n$ is the number of folds. Approximate 95\% confidence intervals can be obtained via $\pm1.96se_{cv}$ The diagonal represents in-dataset validation, i.e., training and test samples were taken from the same dataset. In \textit{pooled (n-1)}, the model was trained on combined data from all except the test dataset. In \textit{all}, training data from all datasets (including the test dataset) was used during model development. AUROC, area under the receiver operating characteristic; ICU, intensive care unit.
        \end{tablenotes}
    \end{threeparttable}
\end{table}

\begin{table}[ht]
    \scriptsize
    \caption{Mean AUROC and standard errors achieved by different network architectures when trained to predict AKI on one dataset (rows) and evaluated on others (columns). }
    \label{tab:erm-aki-detail}
    \centering
    \begin{threeparttable}
        \begin{tabularx}{\textwidth}{Xcccr}
            \toprule
            \textbf{Training domain} & \multicolumn{4}{c}{\textbf{Test domain}} \\\cline{2-5}
             & AUMCdb & HiRID & eICU & MIMIC \\
            \midrule
            \addlinespace[0.3em]
            \multicolumn{5}{l}{\textbf{Gated Recurrent Network}}\\
            AUMCdb & 0.885±0.004 & 0.645±0.006 & 0.675±0.005 & 0.766±0.003 \\ 
            HiRID & 0.704±0.009 & 0.826±0.003 & 0.640±0.006 & 0.641±0.006 \\ 
            eICU & 0.866±0.003 & 0.707±0.008 & 0.875±0.002 & 0.848±0.002 \\ 
            MIMIC & 0.884±0.004 & 0.709±0.003 & 0.842±0.004 & 0.876±0.002 \\ 
            pooled (n-1) & 0.856±0.004 & 0.683±0.008 & 0.804±0.002 & 0.835±0.001 \\ 
            all & 0.899±0.002 & 0.815±0.002 & 0.863±0.005 & 0.873±0.004 \\ 
            \midrule
            \multicolumn{5}{l}{\textbf{Temporal Convolutional Network}}\\
            AUMCdb & 0.887±0.003 & 0.640±0.011 & 0.657±0.005 & 0.724±0.003 \\ 
            HiRID & 0.743±0.017 & 0.824±0.003 & 0.644±0.010 & 0.636±0.011 \\ 
            eICU & 0.889±0.004 & 0.715±0.006 & 0.891±0.002 & 0.868±0.004 \\ 
            MIMIC & 0.907±0.002 & 0.689±0.003 & 0.843±0.005 & 0.888±0.003 \\ 
            pooled (n-1) & 0.883±0.003 & 0.679±0.008 & 0.784±0.007 & 0.850±0.002 \\ 
            all & 0.905±0.003 & 0.815±0.003 & 0.873±0.002 & 0.886±0.002 \\ 
            \midrule
            \multicolumn{5}{l}{\textbf{Transformer}}\\
            AUMCdb & 0.838±0.004 & 0.693±0.004 & 0.612±0.004 & 0.667±0.005 \\ 
            HiRID & 0.733±0.013 & 0.803±0.002 & 0.641±0.003 & 0.638±0.007 \\ 
            eICU & 0.779±0.011 & 0.732±0.006 & 0.782±0.028 & 0.746±0.025 \\ 
            MIMIC & 0.815±0.010 & 0.735±0.006 & 0.641±0.007 & 0.780±0.010 \\ 
            pooled (n-1) & 0.790±0.002 & 0.702±0.014 & 0.651±0.001 & 0.722±0.003 \\ 
            all & 0.815±0.009 & 0.781±0.003 & 0.697±0.003 & 0.750±0.004 \\ 
            \bottomrule
        \end{tabularx}
        \begin{tablenotes}[flushleft, para]
            \footnotesize
            \item \leavevmode\kern-\scriptspace\kern-\labelsep Standard errors were calculated based on 5-fold cross-validation as $se_{cv} = \sigma_{cv}n^{-1/2}$, where $\sigma_{cv}$ is the standard deviation across cross-validation folds and $n$ is the number of folds. Approximate 95\% confidence intervals can be obtained via $\pm1.96se_{cv}$ The diagonal represents in-dataset validation, i.e., training and test samples were taken from the same dataset. In \textit{pooled (n-1)}, the model was trained on combined data from all except the test dataset. In \textit{all}, training data from all datasets (including the test dataset) was used during model development. AKI, acute kidney injury; AUROC, area under the receiver operating characteristic.
        \end{tablenotes}
    \end{threeparttable}
\end{table}

\begin{table}[ht]
    \scriptsize
    \caption{Mean AUROC and standard errors achieved by different network architectures when trained to predict sepsis on one dataset (rows) and evaluated on others (columns). }
    \label{tab:erm-sepsis-detail}
    \centering
    \begin{threeparttable}
        \begin{tabularx}{\textwidth}{Xcccr}
            \toprule
            \textbf{Training domain} & \multicolumn{4}{c}{\textbf{Test domain}} \\\cline{2-5}
             & AUMCdb & HiRID & eICU & MIMIC \\
            \midrule
            \addlinespace[0.3em]
            \multicolumn{5}{l}{\textbf{Gated Recurrent Network}}\\
            AUMCdb & 0.753±0.013 & 0.672±0.006 & 0.598±0.007 & 0.673±0.013 \\ 
            HiRID & 0.723±0.004 & 0.785±0.006 & 0.597±0.013 & 0.657±0.014 \\ 
            eICU & 0.697±0.012 & 0.695±0.007 & 0.741±0.002 & 0.764±0.010 \\ 
            MIMIC & 0.700±0.004 & 0.722±0.003 & 0.697±0.005 & 0.816±0.004 \\ 
            pooled (n-1) & 0.742±0.009 & 0.746±0.003 & 0.711±0.003 & 0.787±0.004 \\ 
            all & 0.808±0.007 & 0.801±0.004 & 0.744±0.002 & 0.831±0.003 \\ 
            \midrule
            \multicolumn{5}{l}{\textbf{Temporal Convolutional Network}}\\
            AUMCdb & 0.780±0.008 & 0.680±0.003 & 0.582±0.004 & 0.677±0.008 \\ 
            HiRID & 0.689±0.022 & 0.795±0.003 & 0.625±0.011 & 0.685±0.012 \\ 
            eICU & 0.709±0.009 & 0.694±0.007 & 0.753±0.004 & 0.771±0.002 \\ 
            MIMIC & 0.728±0.005 & 0.752±0.006 & 0.716±0.005 & 0.824±0.002 \\ 
            pooled (n-1) & 0.764±0.006 & 0.749±0.005 & 0.705±0.005 & 0.784±0.003 \\ 
            all & 0.811±0.008 & 0.809±0.007 & 0.746±0.002 & 0.832±0.003 \\ 
            \midrule
            \multicolumn{5}{l}{\textbf{Transformer}}\\
            AUMCdb & 0.769±0.017 & 0.658±0.004 & 0.586±0.002 & 0.655±0.007 \\ 
            HiRID & 0.694±0.013 & 0.795±0.001 & 0.574±0.012 & 0.634±0.013 \\ 
            eICU & 0.660±0.012 & 0.691±0.008 & 0.736±0.009 & 0.760±0.007 \\ 
            MIMIC & 0.673±0.010 & 0.732±0.004 & 0.690±0.004 & 0.816±0.005 \\ 
            pooled (n-1) & 0.717±0.009 & 0.731±0.009 & 0.694±0.002 & 0.763±0.009 \\ 
            all & 0.805±0.006 & 0.803±0.002 & 0.727±0.005 & 0.817±0.004 \\ 
            \bottomrule
        \end{tabularx}
        \begin{tablenotes}[flushleft, para]
            \footnotesize
            \item \leavevmode\kern-\scriptspace\kern-\labelsep Standard errors were calculated based on 5-fold cross-validation as $se_{cv} = \sigma_{cv}n^{-1/2}$, where $\sigma_{cv}$ is the standard deviation across cross-validation folds and $n$ is the number of folds. Approximate 95\% confidence intervals can be obtained via $\pm1.96se_{cv}$ The diagonal represents in-dataset validation, i.e., training and test samples were taken from the same dataset. In \textit{pooled (n-1)}, the model was trained on combined data from all except the test dataset. In \textit{all}, training data from all datasets (including the test dataset) was used during model development. AKI, acute kidney injury; AUROC, area under the receiver operating characteristic.
        \end{tablenotes}
    \end{threeparttable}
\end{table}

\end{document}